\documentclass[5p]{elsarticle}

\usepackage{graphicx}
\usepackage{caption}
\usepackage{subcaption}
\usepackage{amsmath}
\usepackage{algorithmic}
\usepackage{algorithm}
\usepackage{lineno}
\usepackage{multirow}
\usepackage{tabularx}
\usepackage{booktabs}
\usepackage{url}
\usepackage{color}
\usepackage[figuresright]{rotating}
\usepackage[utf8]{inputenc}

\newcommand{\first}[1]{\textbf{#1}}
\newcommand{\second}[1]{{#1}}

\newcommand{\firstrev}[1]{\textcolor{black}{#1}}

\newcolumntype{Y}{>{\arraybackslash}X}

\journal{Pattern Recognition}

\begin{document}

\begin{frontmatter}

\title{Radial-Based Undersampling for Imbalanced Data Classification}

\author[agh]{Micha\l{} Koziarski\corref{cor1}}
\ead{michal.koziarski@agh.edu.pl}
\cortext[cor1]{Corresponding author}

\address[agh]{Department of Electronics, AGH University of Science and Technology, Al. Mickiewicza 30, 30-059 Krak\'ow, Poland}

\begin{abstract}
Data imbalance remains one of the most widespread problems affecting contemporary machine learning. The negative effect data imbalance can have on the traditional learning algorithms is most severe in combination with other dataset difficulty factors, such as small disjuncts, presence of outliers and insufficient number of training observations. \firstrev{Aforementioned difficulty factors can also limit the applicability of some of the methods of dealing with data imbalance, in particular the neighborhood-based oversampling algorithms based on SMOTE.} Radial-Based Oversampling (RBO) was previously proposed to mitigate some of the limitations of the neighborhood-based methods. In this paper we examine the possibility of utilizing the concept of mutual class potential, used to guide the oversampling process in RBO, in the undersampling procedure. Conducted computational complexity analysis indicates a significantly reduced time complexity of the proposed Radial-Based Undersampling algorithm, and the results of the performed experimental study indicate its usefulness, especially on difficult datasets.
\end{abstract}


\end{frontmatter}


\section{Introduction}

The problem of data imbalance \cite{Sun:2009,Krawczyk:2016,Branco:2016} occurs in the classification task whenever the number of observations belonging to one of the classes, the \textit{majority class}, exceeds the number of observations belonging to one of the other classes, the \textit{minority class}. Traditional classification algorithms are susceptible to the presence of imbalanced data, and tend to display a bias towards the majority class at the expense of the capability of minority class discrimination. This negative effect on the classification performance is further exacerbated by a presence of additional dataset difficulty factors, such as small disjuncts \cite{jo2004class} or insufficient number of training observations \cite{chen2008fast}, that can lead to model overfitting.

Most real datasets exhibit some degree of imbalance that can influence the classification process. Data imbalance heavily impacts many practical domains, such as cancer malignancy grading \cite{krawczyk2016evolutionary,koziarski2018convolutional}, industrial systems monitoring \cite{ramentol2016fuzzy}, fraud detection \cite{wei2013effective}, behavioral analysis \cite{azaria2014behavioral} and cheminformatics \cite{czarnecki2015compounds}. As a result, imbalanced data classification remains an active area of research \cite{fernandez2017pareto,koziarski2017ccr,lango2018multi,ksieniewicz2018imbalanced}. Numerous approaches to mitigating the negative impact of data imbalance have been proposed in the literature. In particular, a family of data-level methods can be distinguished. Data-level methods manipulate with the training data to make it more suitable for classification by traditional learning algorithms, either by increasing the number of minority observations (oversampling) or reducing the number of majority observations (undersampling).

\firstrev{Perhaps the most widespread paradigm of imbalanced data resampling are the neighborhood-based algorithms based on Synthetic Minority Oversampling Technique (SMOTE) \cite{chawla2002smote}}. However, SMOTE and many of its derivatives are susceptible to the presence of difficulty factors such as small disjuncts, outliers and small number of minority observations. Recently, a novel method based on the concept of mutual class potential, Radial-Based Oversampling (RBO) \cite{koziarski2017radial}, has been proposed with the goal of avoiding some of the pitfalls of SMOTE.

In this paper we investigate the possibility of extending the concept of mutual class potential to the undersampling procedure, with the aim of preserving some of the performance gains offered by using the potential to guide the resampling, while simultaneously reducing the computational complexity of the algorithm. \firstrev{The contributions of this paper can be summarized as follows.
\begin{enumerate}
\item Proposal of a novel Radial-Based Undersampling (RBU) algorithm based on the concept of mutual class potential.
\item Detailed analysis of the computational complexity of the proposed method, indicating a significantly reduced cost compared to the RBO algorithm.
\item Conceptual and experimental analysis of the impact of the RBUs parameters on its behavior and resulting performance.
\item Thorough experimental evaluation of the proposed method on the basis of diverse benchmark datasets and a large number of state-of-the-art, data-level reference algorithms.
\item Analysis of the impact of dataset characteristics on the relative performance of the proposed method.
\end{enumerate}
In other words, the paper expands on the idea of the mutual class potential and proposes a significantly faster alternative to the RBO algorithm, directly addressing one of the main drawbacks of the original oversampling approach. At the same, during the conducted experimental analysis RBU achieved performance comparable to the original RBO algorithm, with a statistically significantly better results when combined with the decision trees. Finally, presented conceptual and experimental analysis provide some insight on the strengths, weaknesses and areas of applicability of the proposed approach.
}

The rest of this paper is organized as follows. \firstrev{In Section~\ref{sec:related} we present the related work on the neighborhood-based oversampling algorithms, guided undersampling techniques, advantages and disadvantages of the two, and the categorization of the minority object types.} In Section~\ref{sec:rbu} we describe the proposed method and discuss its computational complexity. In Section~\ref{sec:exp} we describe the conducted experimental study and the observed results. Finally, in Section~\ref{sec:con} we present our conclusions.

\section{Related Work}
\label{sec:related}

In this section we discuss the research relevant to the approach proposed in this paper. We begin with a brief description of the most prevalent paradigm of imbalanced data oversampling, neighborhood-based methods, and highlight their shortcomings which originally inspired Radial-Based Oversampling. Afterwards we describe the related research on guided undersampling strategies and briefly outline the most notable algorithms. \firstrev{Next we discuss the advantages and disadvantages of applying either over- or undersampling, and present some of the considerations that may affect the choice of the resampling strategy.} Finally, we summarize the existing research of minority object type categorization, used later in this paper to identify the areas of applicability of the proposed method.

\subsection{Neighborhood-based oversampling algorithms}

The most fundamental choice during the design of both oversampling and undersampling algorithms for handling data imbalance is the question of defining the regions of interest: the areas in which either the new instances are to be placed, in case of oversampling, or from which the existing instances are to be removed, in case of undersampling. Besides the random approaches, probably the most prevalent paradigm for the oversampling are the neighborhood-based methods originating from Synthetic Minority Over-sampling Technique (SMOTE) \cite{chawla2002smote}. The regions of interest of SMOTE are located between any given minority observation and its closest minority neighbors: SMOTE synthesizes new instances by interpolating the observation and one of its, randomly selected, nearest neighbors.

\begin{figure}
\centering
\includegraphics[width=0.49\linewidth]{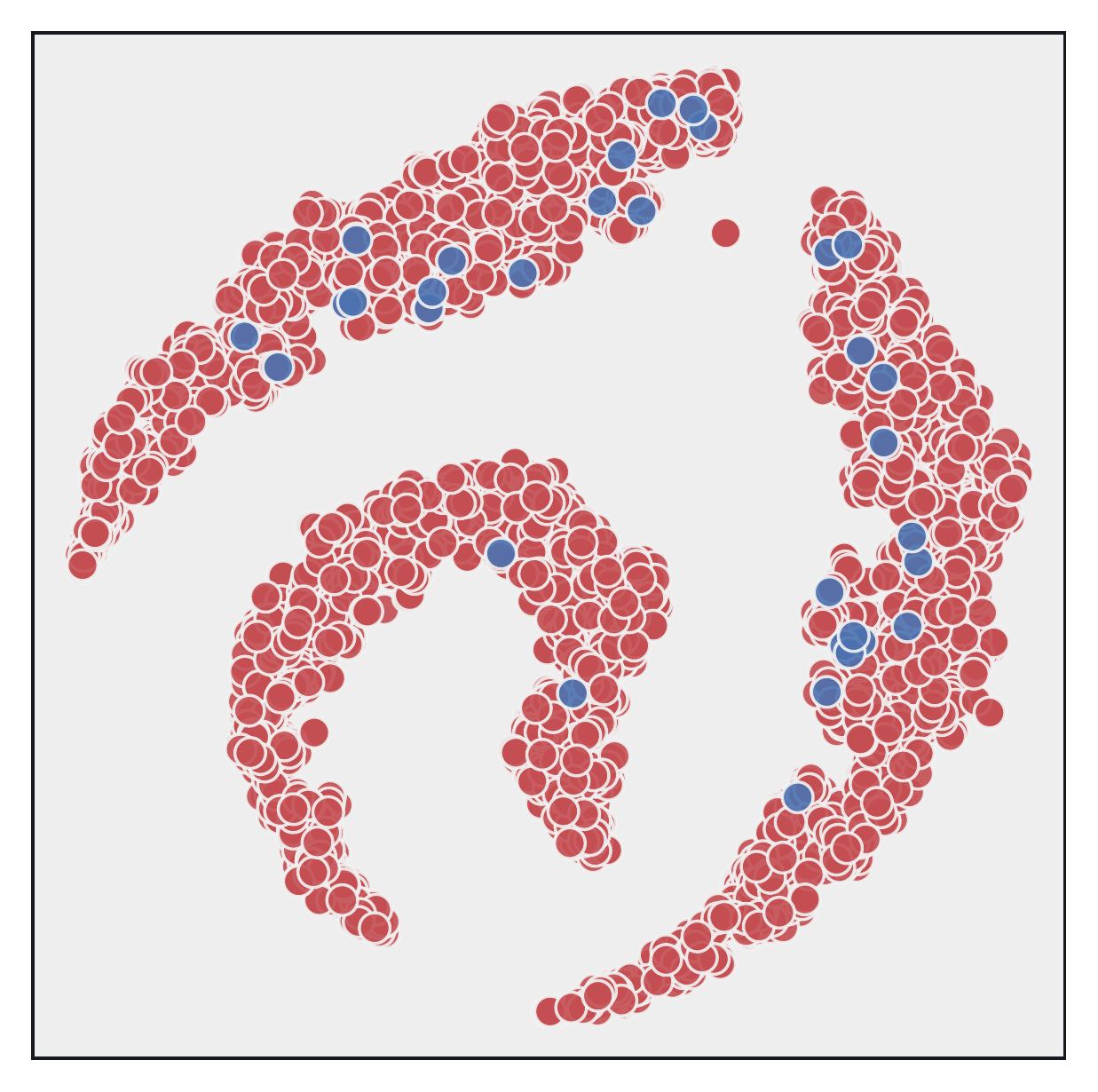}
\includegraphics[width=0.49\linewidth]{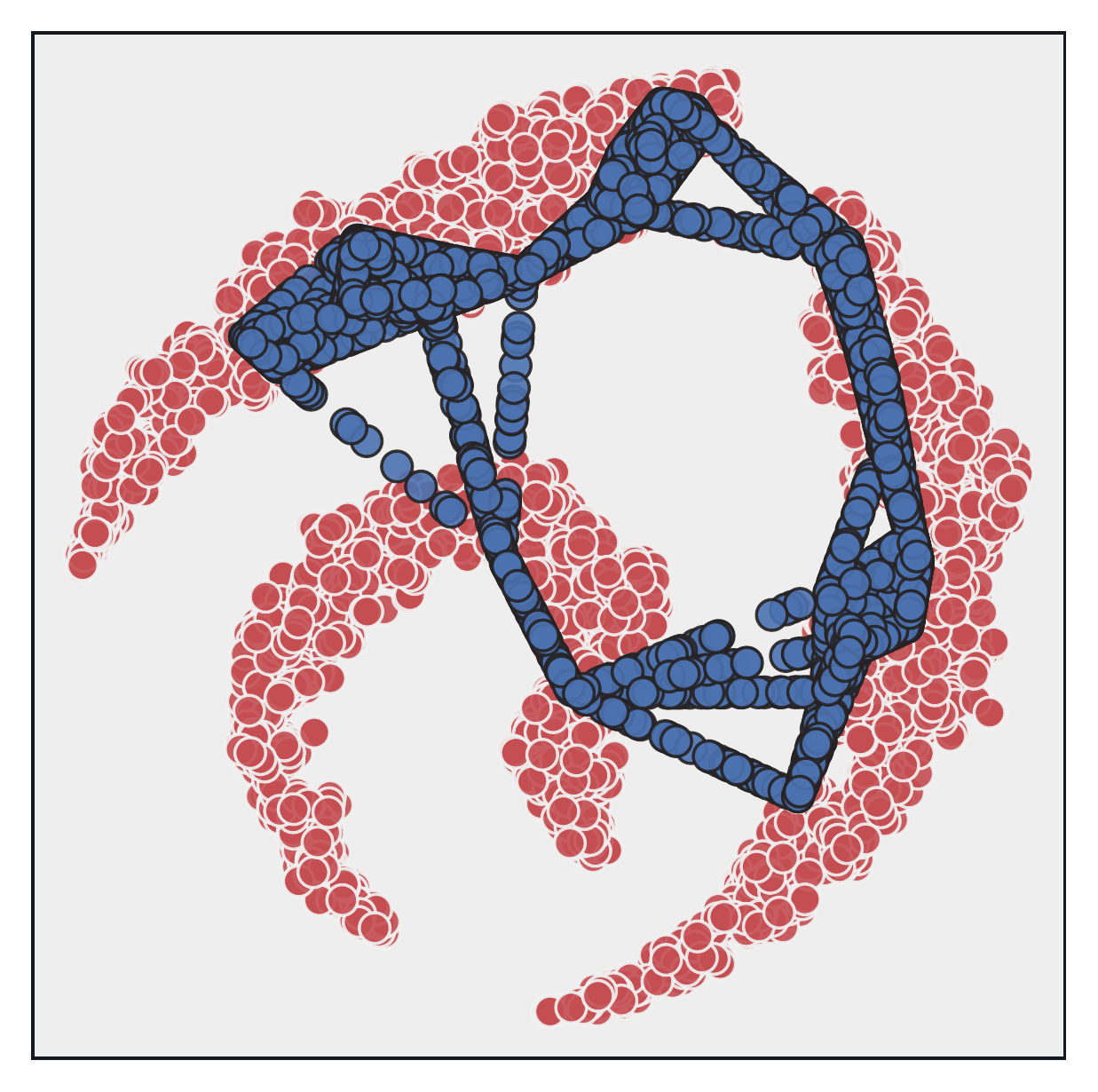}
\caption{An example of a difficult dataset for neighborhood-based methods, displaying factors such as a small number of minority objects, disjoint data distributions, or presence of the outliers. On the left: original data distribution. On the right: dataset after oversampling with SMOTE, with the generated observations highlighted.}
\label{fig:example-smote}
\end{figure}

SMOTE can be considered a cornerstone for the majority of the existing oversampling strategies \cite{perez2016oversampling,bellinger2018manifold}. Numerous extensions of the original algorithm were proposed, with the most notable including: Borderline-SMOTE \cite{han2005borderline}, focusing on the borderline instances, placed close to the decision border; Adaptive Synthetic Sampling (ADASYN) \cite{he2008adasyn}, individually adjusting the oversampling ratio based on the difficulty of the given observation; and Safe-Level-SMOTE \cite{Bunkhumpornpat:2009} and LN-SMOTE \cite{Maciejewski:2011}, limiting the risk of placing synthetic instances inside the regions belonging to the majority class. However, despite their prevalence, neighborhood-based approaches have their own shortcomings that can affect the suitability of synthesized observations for improving classification. Most importantly, in the basic variant SMOTE does not utilize the information about the distribution of the majority class objects: the regions of interest are based solely on the position of the minority observations. This can potentially lead to synthesizing minority observations overlapping the clusters of majority observations for datasets displaying factors such as a small number of minority objects, disjoint data distributions, or presence of the outliers, which was illustrated in Figure~\ref{fig:example-smote}. Classifiers trained on datasets resampled in that way can display, possibly unjustified, bias towards the minority class and resulting decreased performance. While some attempts at limiting the described deficiency of SMOTE have been made, such as the previously mentioned extensions, Safe-Level-SMOTE and LN-SMOTE, or combining the oversampling using SMOTE with later cleaning with either Tomek links \cite{tomek1976two} or Edited Nearest-Neighbor rule \cite{wilson1972asymptotic}, a further research into the methods explicitly using the information about the distribution of both classes is required.

\subsection{Guided undersampling strategies}

Similar to the case of oversampling, finding the regions of interest, in the case of undersampling indicating which observations are to be discarded, is essential choice in the algorithm design process. Besides the random methods, over the years a number of guided undersampling strategies was proposed. Many of them rely on some sort of mechanism for identifying the least informative instances, either due to a high redundancy of the given observation or a low confidence that it is not an outlier. 

One of the oldest examples of the latter are the cleaning strategies, heuristics algorithms used to remove observations deemed as inconsistent with the remainder of the data: Tomek links \cite{tomek1976two}, Edited Nearest-Neighbor rule \cite{wilson1972asymptotic}, Condensed Nearest Neighbour editing (CNN) \cite{hart1968condensed}, and more recently Near Miss method (NM) \cite{mani2003knn}, constitute examples of that paradigm. Notably, these methods do not allow specifying the number of observations that should be discarded: instead, they remove all the observations meeting the cleaning rule, which can leave an undesired level of data imbalance. As a result, more recent methods tend to sort the majority observations according to the chosen criterion and allow arbitrary level of balancing. For instance, Anand et al. \cite{anand2010approach} propose sorting the undersampled observations based on the weighted Euclidean distance from the positive samples. Smith et al. \cite{smith2014instance}, in their study of instance level data complexity, advocate for using the instance hardness criterion, with the hardness estimated based on the certainty of the classifiers predictions. 

Another family of methods that can be distinguished are the cluster-based undersampling algorithms, notably the methods proposed by Yen and Lee \cite{yen2009cluster}, which use clustering to select the most representative subset of data. Finally, as has been originally demonstrated by Liu et al. \cite{liu2008exploratory}, undersampling algorithms are well-suited for forming classifier ensembles, an idea that was further extended in form of evolutionary undersampling \cite{galar2013eusboost} and boosting \cite{lu2017adaptive}.

\firstrev{
\subsection{Differences between oversampling and undersampling}
}

\firstrev{
On the most basic level both oversampling and undersampling strategies modify the original data distribution to alleviate the problem of imbalance. However, both approaches can lead to a significantly different performance on any given dataset, and pose unique challenges during the algorithm design process. The main issue associated with undersampling is the possibility of discarding valuable information, whereas during the oversampling we are concerned with the possibility of overfitting the classifier and an increased computational cost of training for an artificially enlarged dataset \cite{barandela2004imbalanced}. The resampling process is also vastly different: while during the undersampling we are concerned only with designating a subset of majority observations that should be removed, effectively ranking a finite number of discrete objects, starting with SMOTE oversampling strategies tend to be a continuous problems, requiring finding regions in which synthetic observations should be generated.
}

\firstrev{
The choice of one of these approaches is determined by a number of factors and has been, to some extent, studied in the literature. First of all, some of the classification algorithms show clear preference towards either of the resampling strategies, with a notable example of decision trees, the overfitting of which was a motivation behind the SMOTE \cite{chawla2002smote}. Nevertheless, later study found the SMOTE itself to still be ineffective when combined with the C4.5 algorithm \cite{drummond2003c4}, for which applying undersampling led to a better performance. In another study \cite{van2009knowledge} authors focused on the impact of noise, with a conclusion that especially for a high levels of noise simple random undersampling produced the best results. Finally, in another experimental study \cite{garcia2012effectiveness} authors investigated the impact of the level of imbalance on the choice of the resampling strategy. Their results indicate that oversampling tends to perform better on a severely imbalanced datasets, while for more modest levels of imbalance both over- and undersampling tend to perform similarly. In general, there is no clear consensus on which of the approaches tends to produce the better results, and while some guidelines are available, in practice an experimental evaluation of both approaches is usually required.
}

\subsection{Categorization of minority object types}

Despite the abundance of different strategies of dealing with data imbalance, it often remains unclear under what conditions a given method is expected to guarantee a satisfactory performance. Furthermore, taking into the account the no free lunch theorem \cite{wolpert1996lack} it is unreasonable to expect that any single method will be able to achieve a state-of-the-art performance on every provided dataset. Identifying the areas of applicability, conditions under which the method is expected to be more likely to achieve a good performance, is therefore desirable both from the point of view of a practitioner, who can use that information to narrow down the range of methods appropriate for a problem at hand, as well as a theoretician, who can use that insight in the process of developing novel methods.

In the context of the imbalanced data classification, one of the criteria that can influence the applicability of different resampling strategies are the characteristics of the minority class distribution. Napierała and Stefanowski \cite{napierala2016types} proposed a method of categorization of different types of minority objects that capture these characteristics. Their approach uses a 5-neighborhood to identify the nearest neighbors of a given object, and afterwards assigns to it a category based on the proportion of neighbors from the same class: \textit{safe} in case of 4 or 5 neighbors from the same class, \textit{borderline} in case of 2 to 3 neighbors, \textit{rare} in case of 1 neighbor, and \textit{outlier} when there are no neighbors from the same class. The percentage of the minority objects from different categories can be then used to describe the character of the entire dataset: an example of datasets with a large proportion of different minority object types was presented in Figure~\ref{fig:type}. Note that the imbalance ratio of the dataset does not determine the type of the minority objects it consists of, which was demonstrated in the above example.

\begin{figure}
\centering
\includegraphics[width=0.49\linewidth]{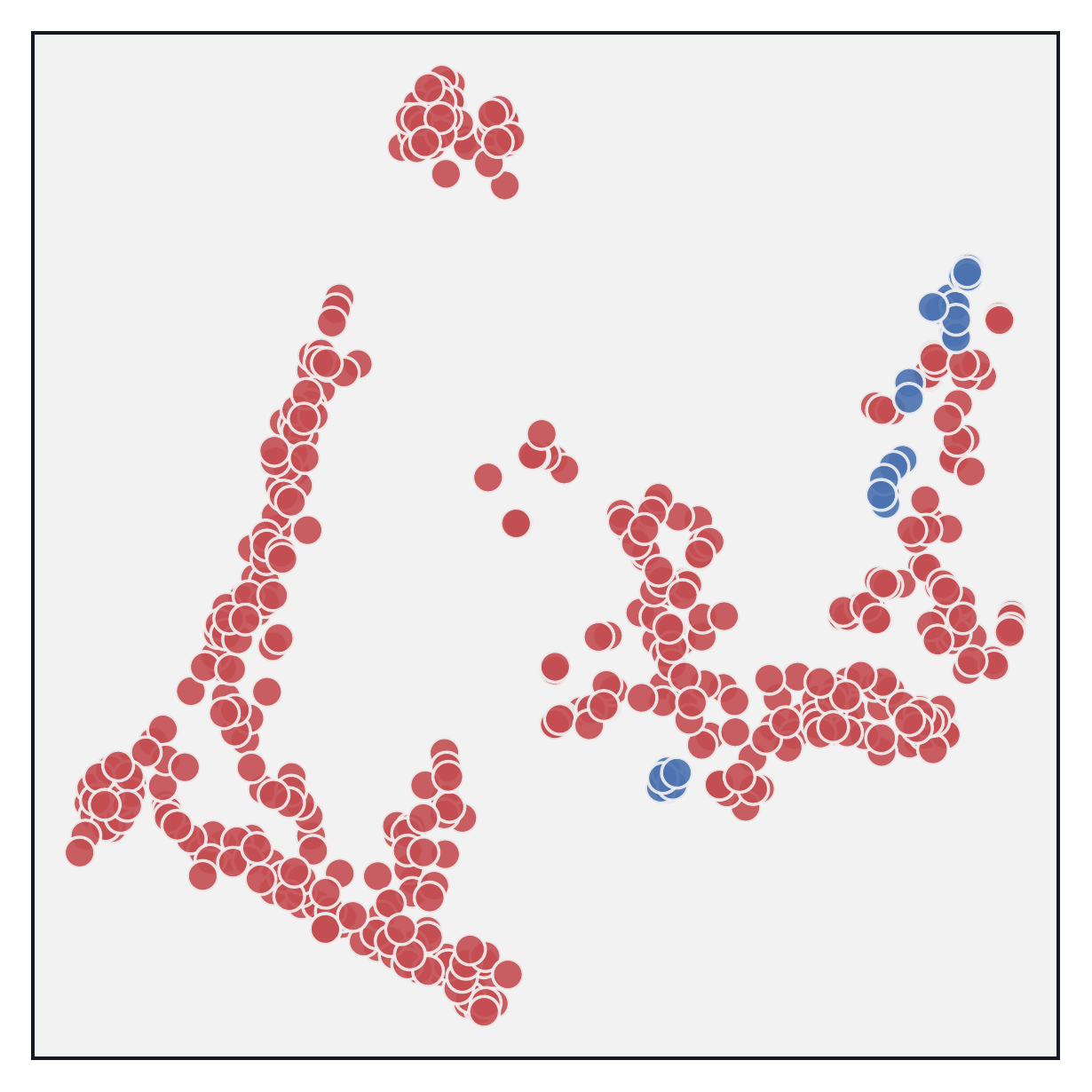}
\includegraphics[width=0.49\linewidth]{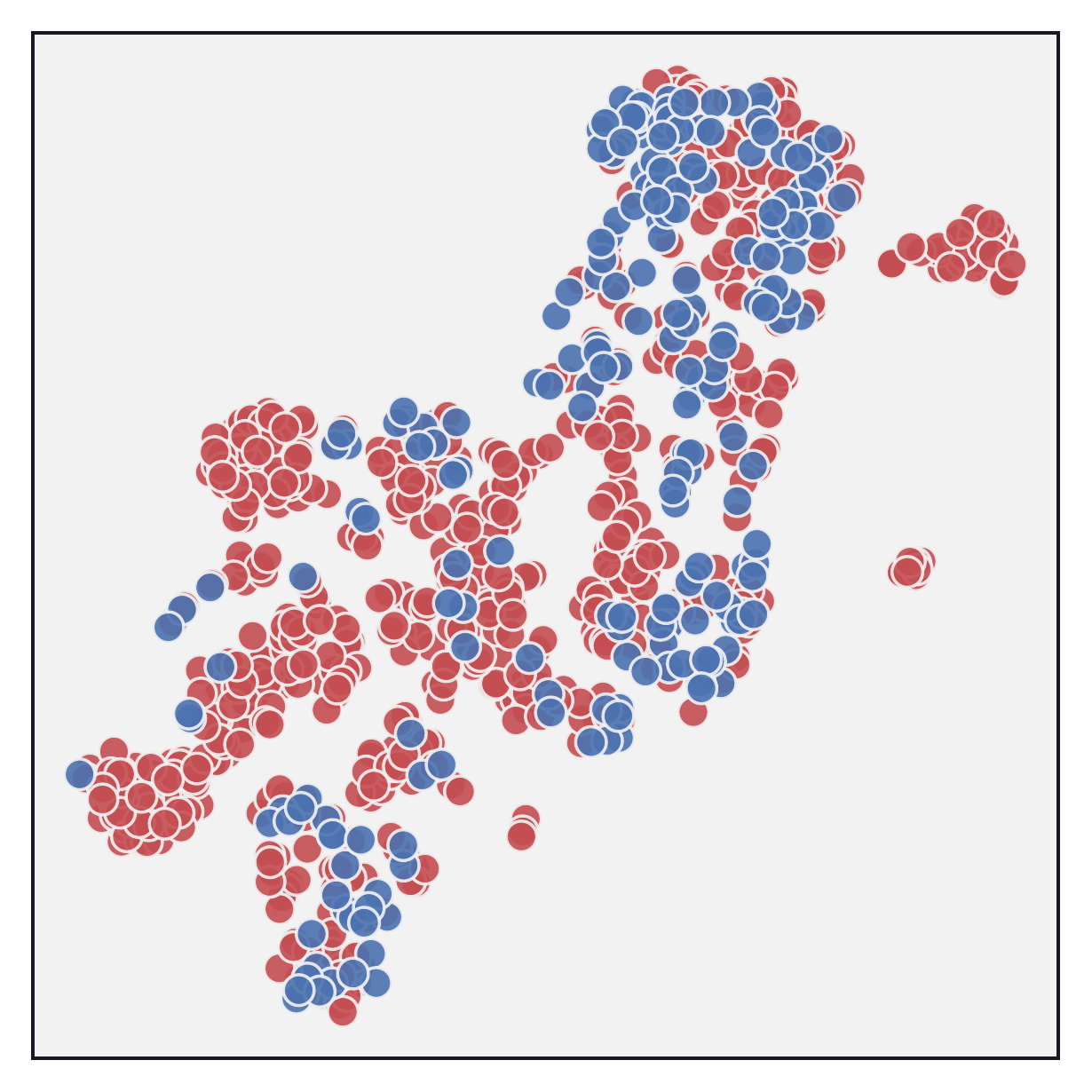}
\includegraphics[width=0.49\linewidth]{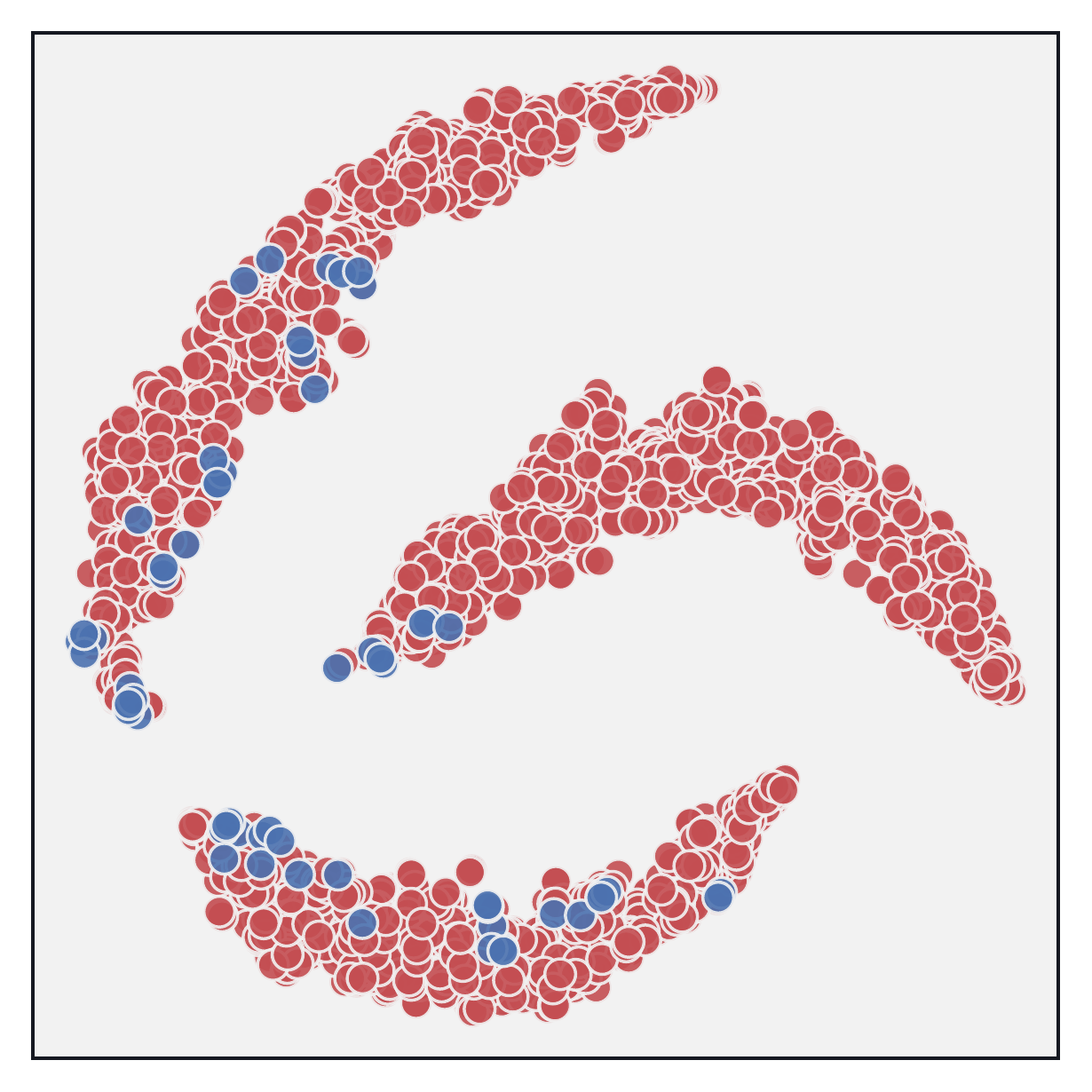}
\includegraphics[width=0.49\linewidth]{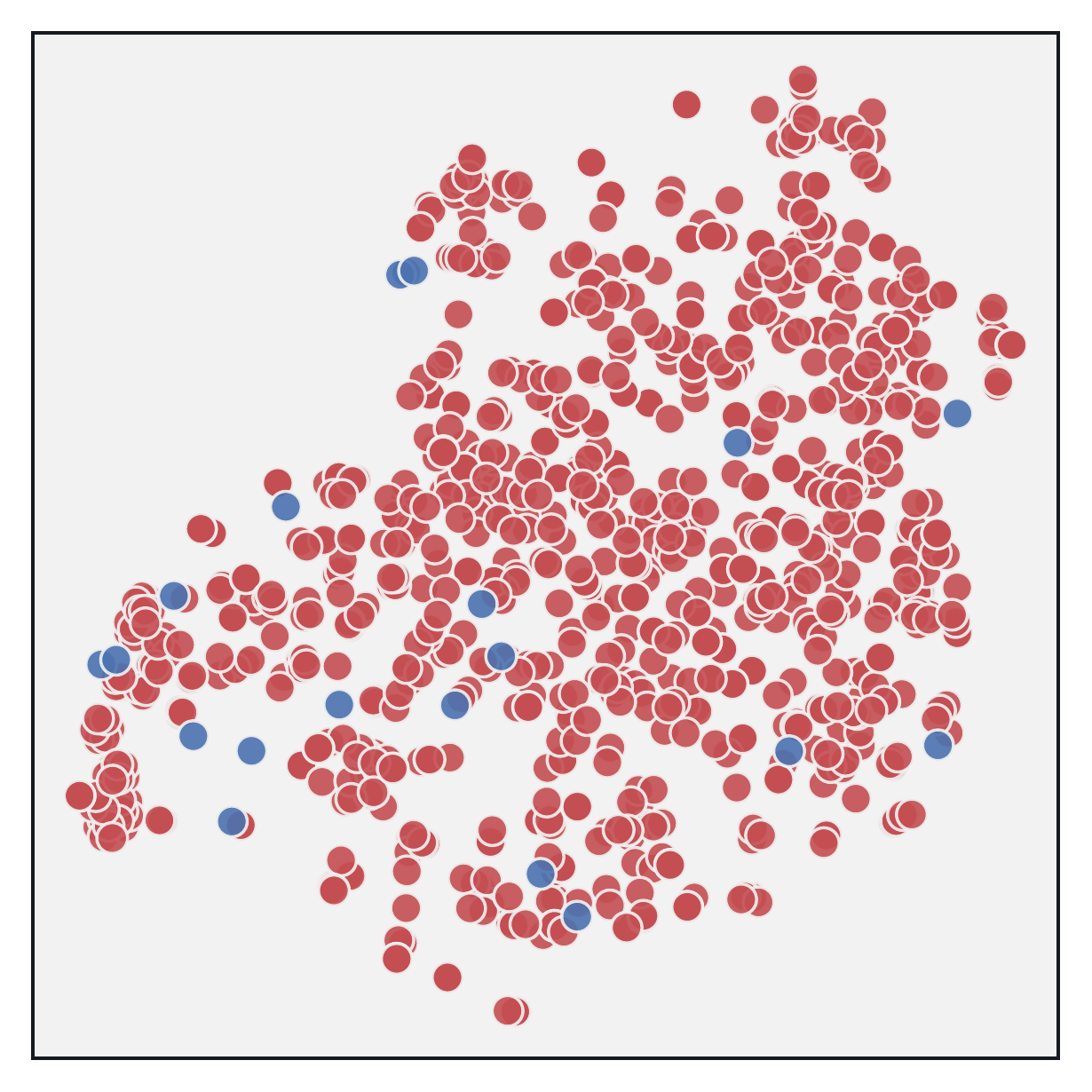}
\caption{An example of datasets with a large proportion of safe (top left), borderline (top right), rare (bottom left) and outlier (bottom right) minority objects.}
\label{fig:example-imba}
\end{figure}

\section{Radial-Based Undersampling}
\label{sec:rbu}

In this section we describe the proposed Radial-Based Undersampling algorithm. We begin with a description of the potential estimation using radial basis functions, and the previously introduced Radial-Based Oversampling algorithm. Afterwards, we describe how the concept of mutual class potential can be applied during the undersampling of the majority objects. Finally, we discuss the computational complexity of the proposed algorithm.

\subsection{Potential estimation and Radial-Based Oversampling}

In the context of imbalanced data resampling, the concept of class potential was first introduced as an approach for designating the regions of interest for oversampling \cite{koziarski2017radial}. Specifically, it was proposed as an alternative to the regions of interest used by SMOTE and its derivatives, which did not utilize the information about the majority class distribution. Mutual class potential is a real-valued function, the value of which, in any given point in space, represents the degree of affiliation of that point to either the majority or the minority class. To calculate that potential, we assign a Gaussian radial basis function (RBF) to every observation in the considered dataset, with the polarity dependent on its class. We assume the convention of assigning a positive polarity to the majority class observations and a negative polarity to the minority class observations. More formally, given a set of majority observations $K$, a set of minority observations $\kappa$, and a parameter $\gamma$ affecting the spread of a single RBF, we define the mutual class potential in the point $x$ as
\begin{linenomath*}
\begin{equation}
\label{eq:potential}
\Phi(x, K, \kappa, \gamma) = \sum_{i=1}^{\mid K \mid}{e^{-(\frac{\lVert K_i - x \rVert_2}{\gamma})^{2}}} - \sum_{j=1}^{\mid \kappa \mid}{e^{-(\frac{\lVert \kappa_j - x \rVert_2}{\gamma})^{2}}},
\end{equation}
\end{linenomath*}
\noindent where $K_i$ denotes the $i$-th object from the majority class and $\kappa_j$ denotes $j$-th object from the minority class, respectively.

Mutual class potential was used in the Radial-Based Oversampling algorithm, in which an iterative optimization was harnessed to locate the regions minimizing the absolute value of potential. Intuitively, such regions represent a low certainty towards the affiliation to either of the classes. New synthetic observations were generated in such regions with the aim of reducing the classifiers bias towards the majority class and moving the decision border in favor of the minority class. Compared with SMOTE, such approach displayed some beneficial characteristics. RBO tended to be less affected by the presence of the outliers, as well as a small number of minority objects combined with disjoint distributions. While in those cases SMOTE was likely to generate new instances overlapping the clusters of existing majority objects, using RBO resulted in a smaller, constrained regions of negative class potential in which new instances were synthesized.

The algorithm was afterwards extended to the problem of classification of noisy imbalanced data \cite{koziarski2019radial} and multi-class imbalanced data \cite{koziarski2019mcrbo}. Furthermore, an extension omitting observations categorized as outliers was also proposed by Bobowska and Woźniak \cite{bobowska2018experimental}. Despite leading to a favorable performance in the conducted experiments, especially in the multi-class setting, using RBO was computationally expensive due to the need of recalculating the class potential at every optimization step. Reducing the computational overhead of the algorithm was an issue identified to be essential to make the algorithm applicable to a very large datasets.

\subsection{Using mutual class potential during undersampling}

While originally proposed to provide the regions of interest in the process of oversampling, mutual class potential can also easily be used to guide the process of undersampling the majority class. Recall that, using the assumed convention, high value of mutual class potential in a given point in space would indicate that in its proximity there is a higher concentration of majority than minority observations. It is therefore possible to rank the existing majority observations based on their mutual class potential. We propose using such ranking mechanism to determine the order of undersampling. Specifically, we make the assumption that the majority observations with highest corresponding mutual class potential provide the least amount of information and are more redundant than the observations with lower potential. As a result, we undersample in the order of decreasing potential, updating its value for the remaining observations after each undersampled object.

\begin{figure*}
\centering
\includegraphics[width=0.195\textwidth]{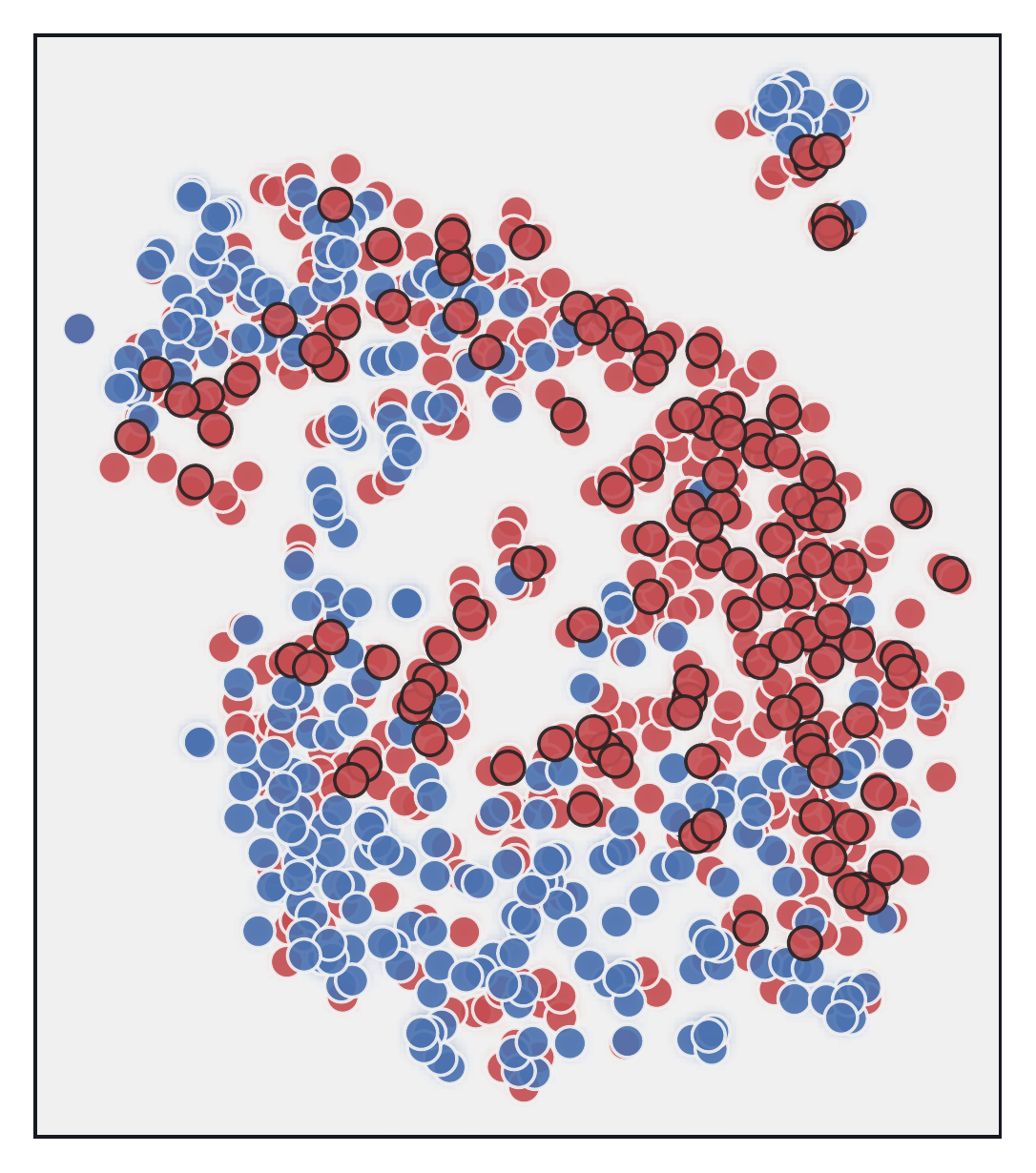}
\includegraphics[width=0.195\textwidth]{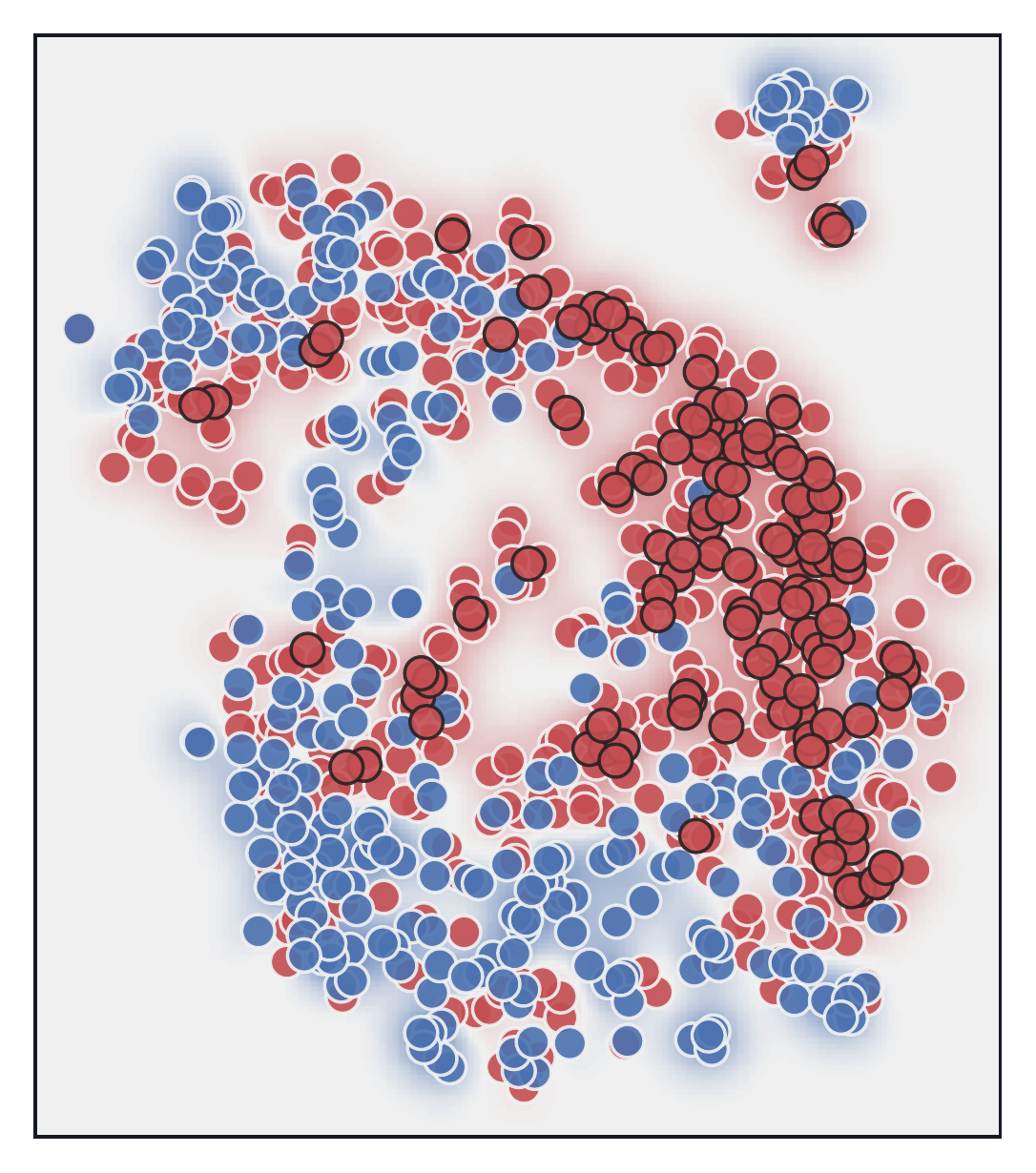}
\includegraphics[width=0.195\textwidth]{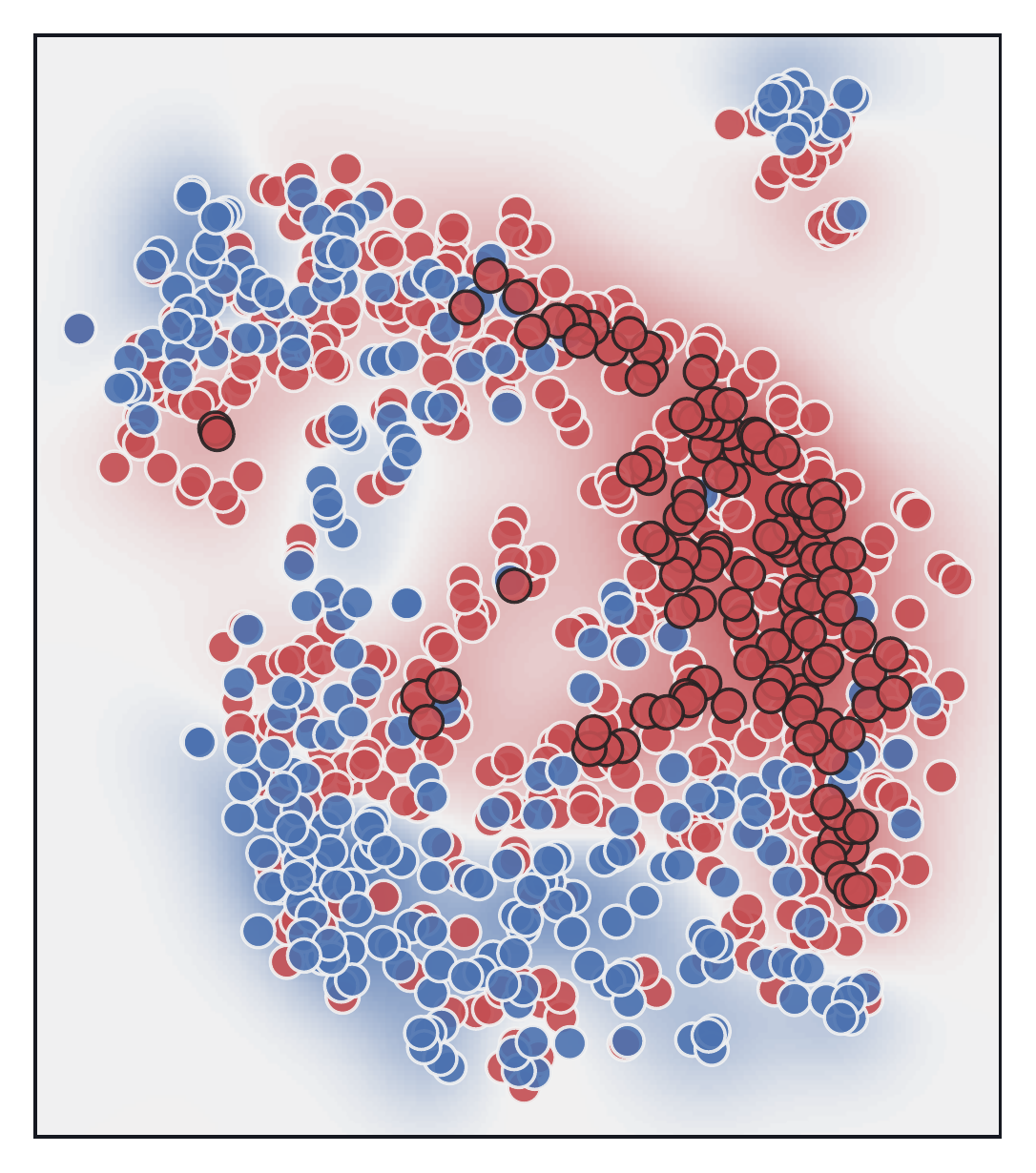}
\includegraphics[width=0.195\textwidth]{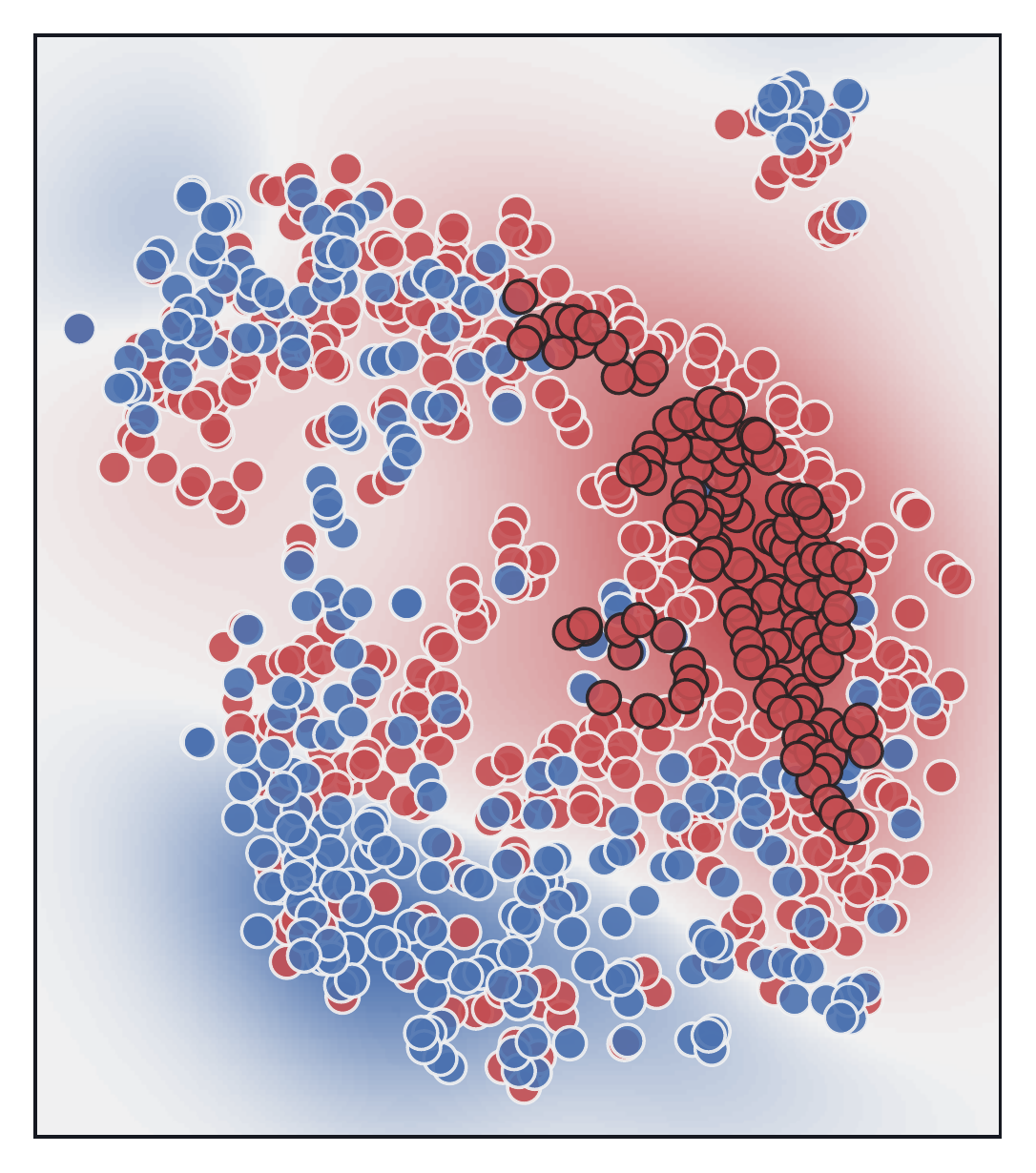}
\includegraphics[width=0.195\textwidth]{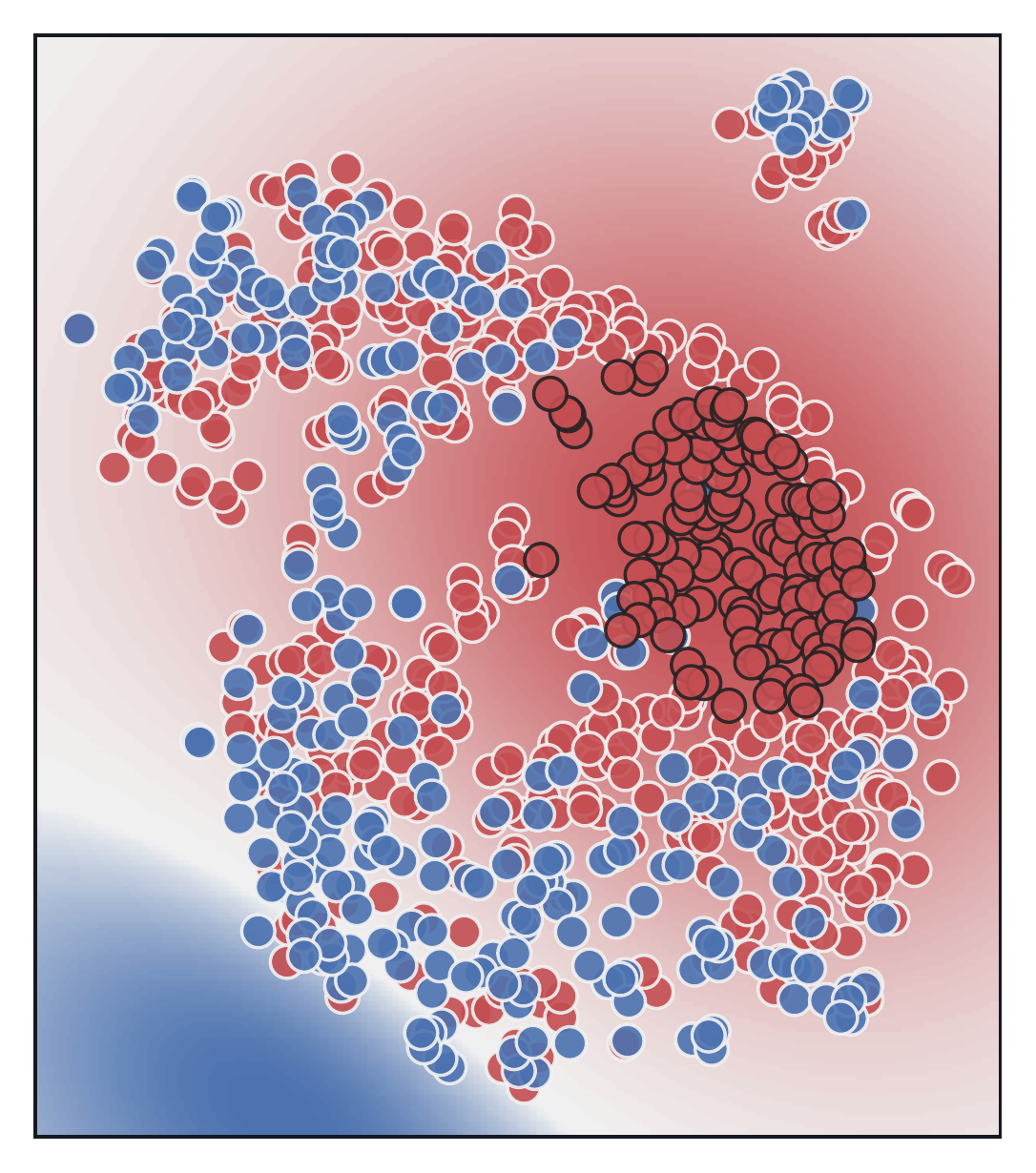}

\includegraphics[width=0.195\textwidth]{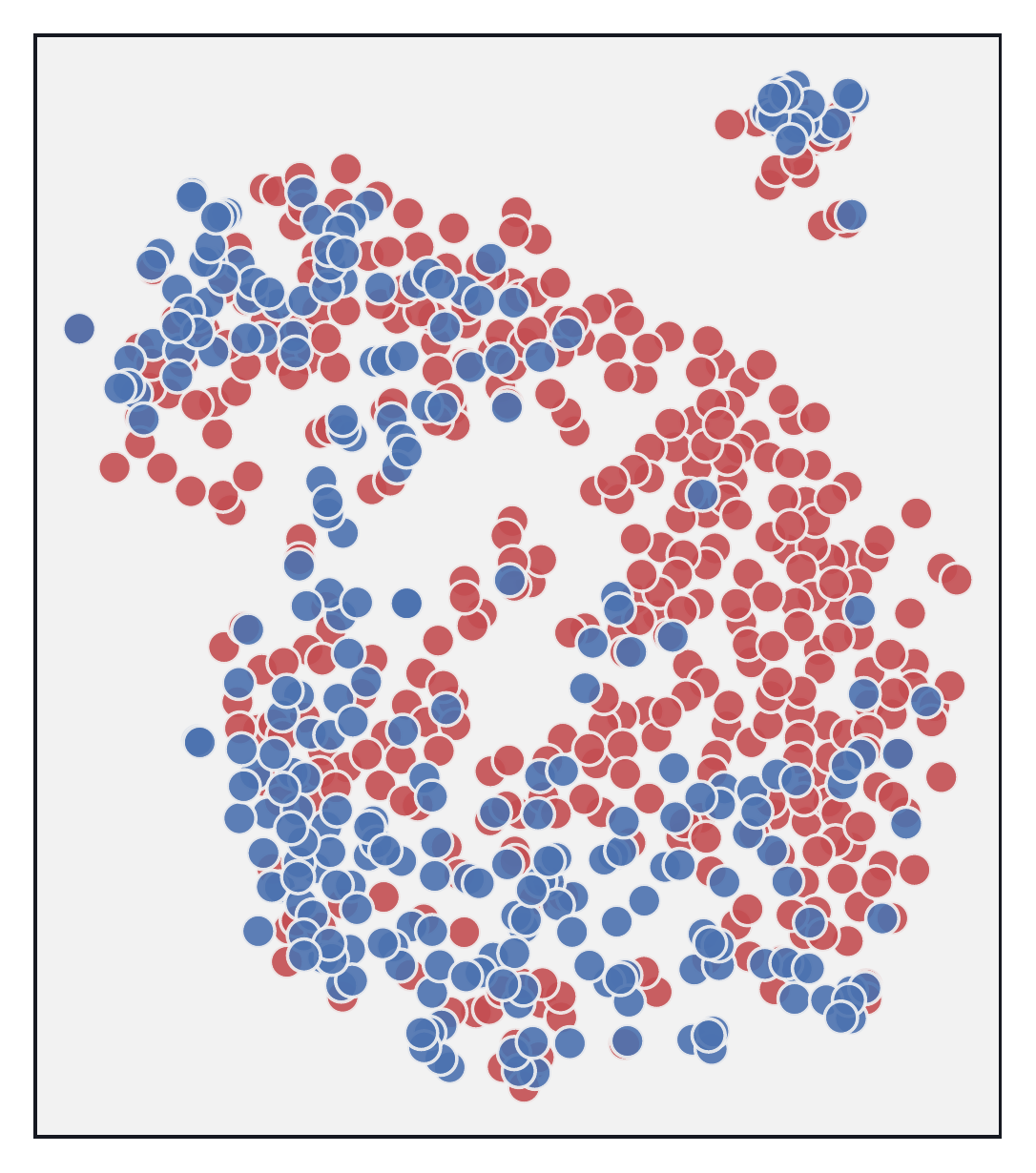}
\includegraphics[width=0.195\textwidth]{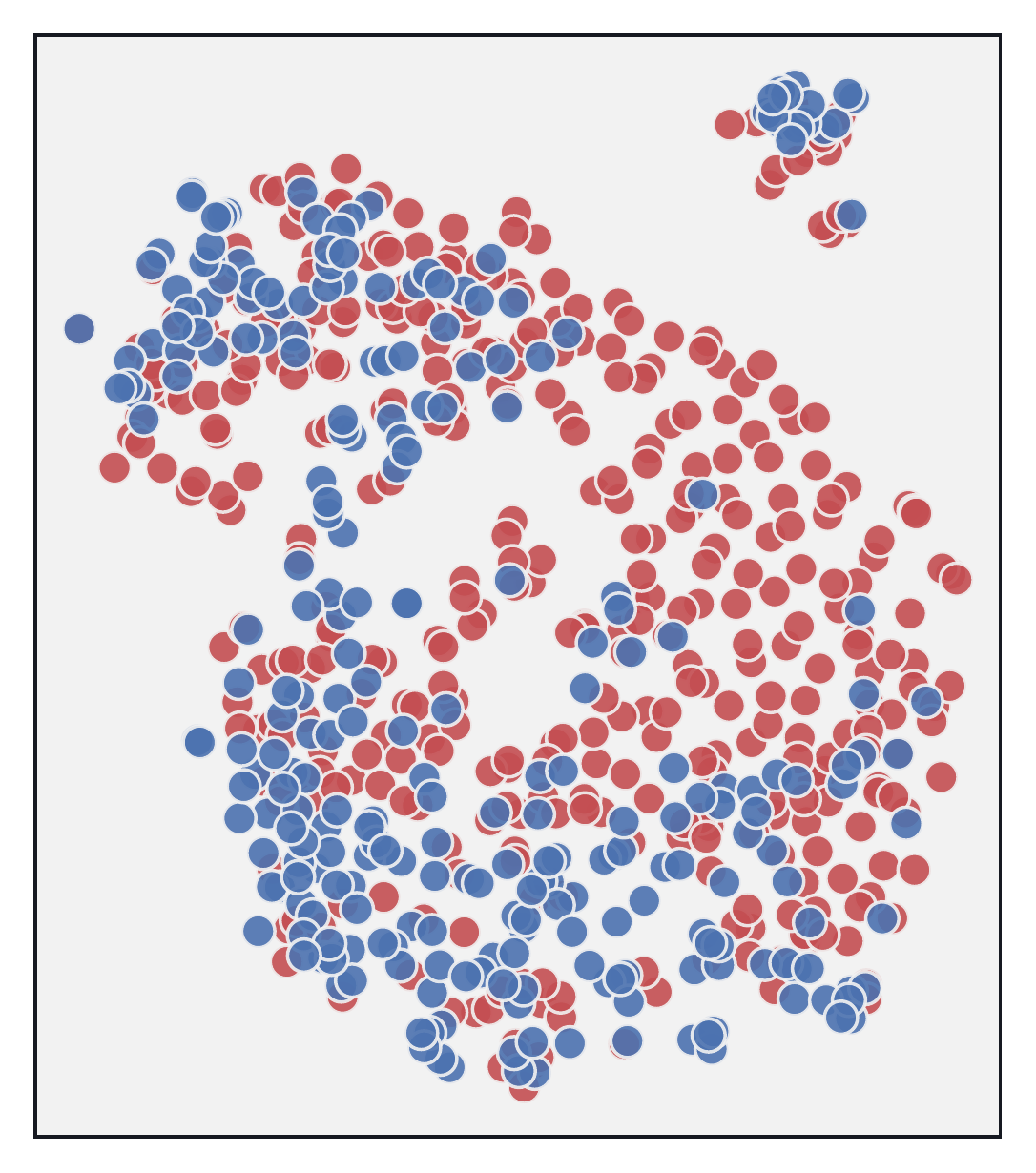}
\includegraphics[width=0.195\textwidth]{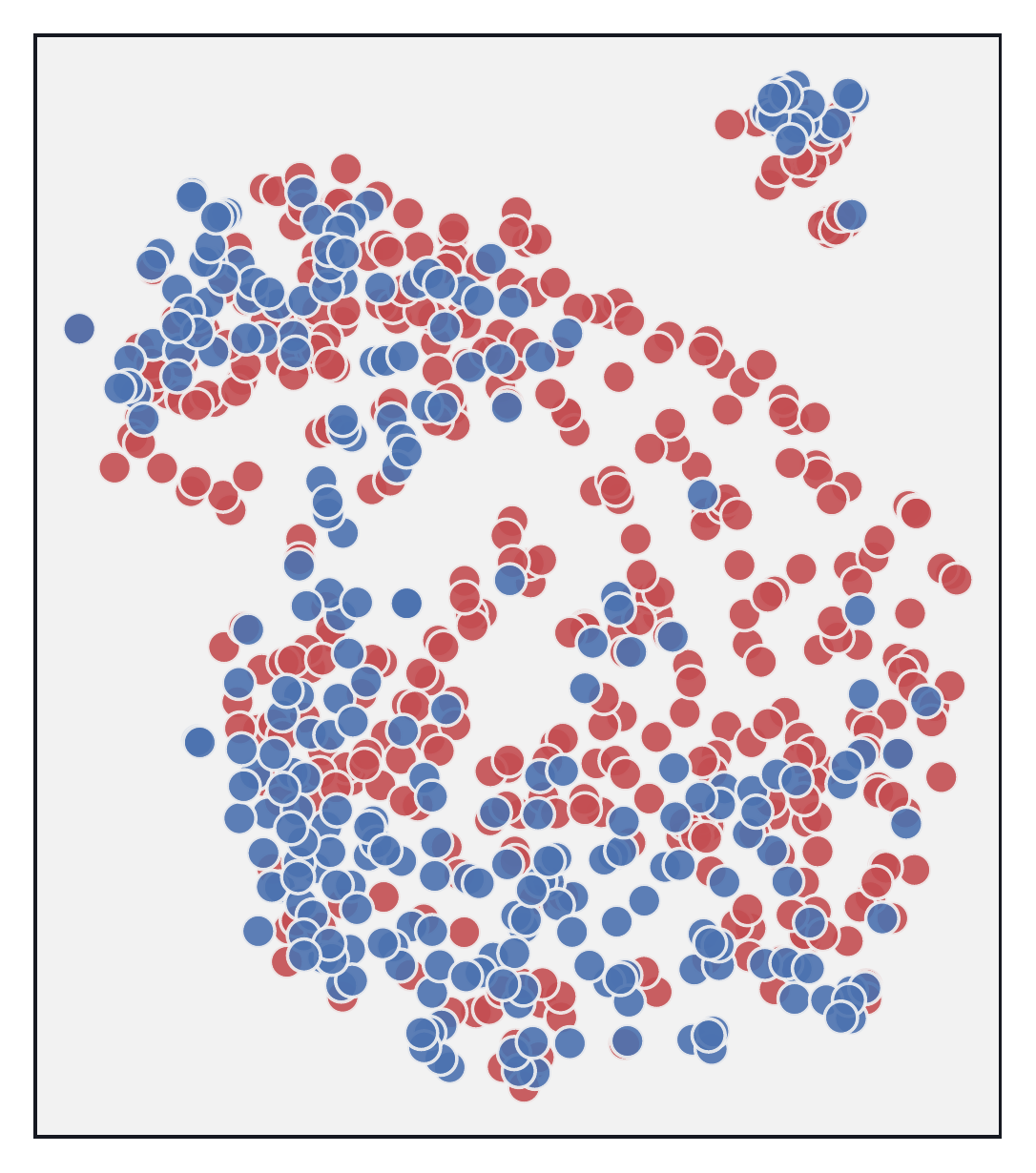}
\includegraphics[width=0.195\textwidth]{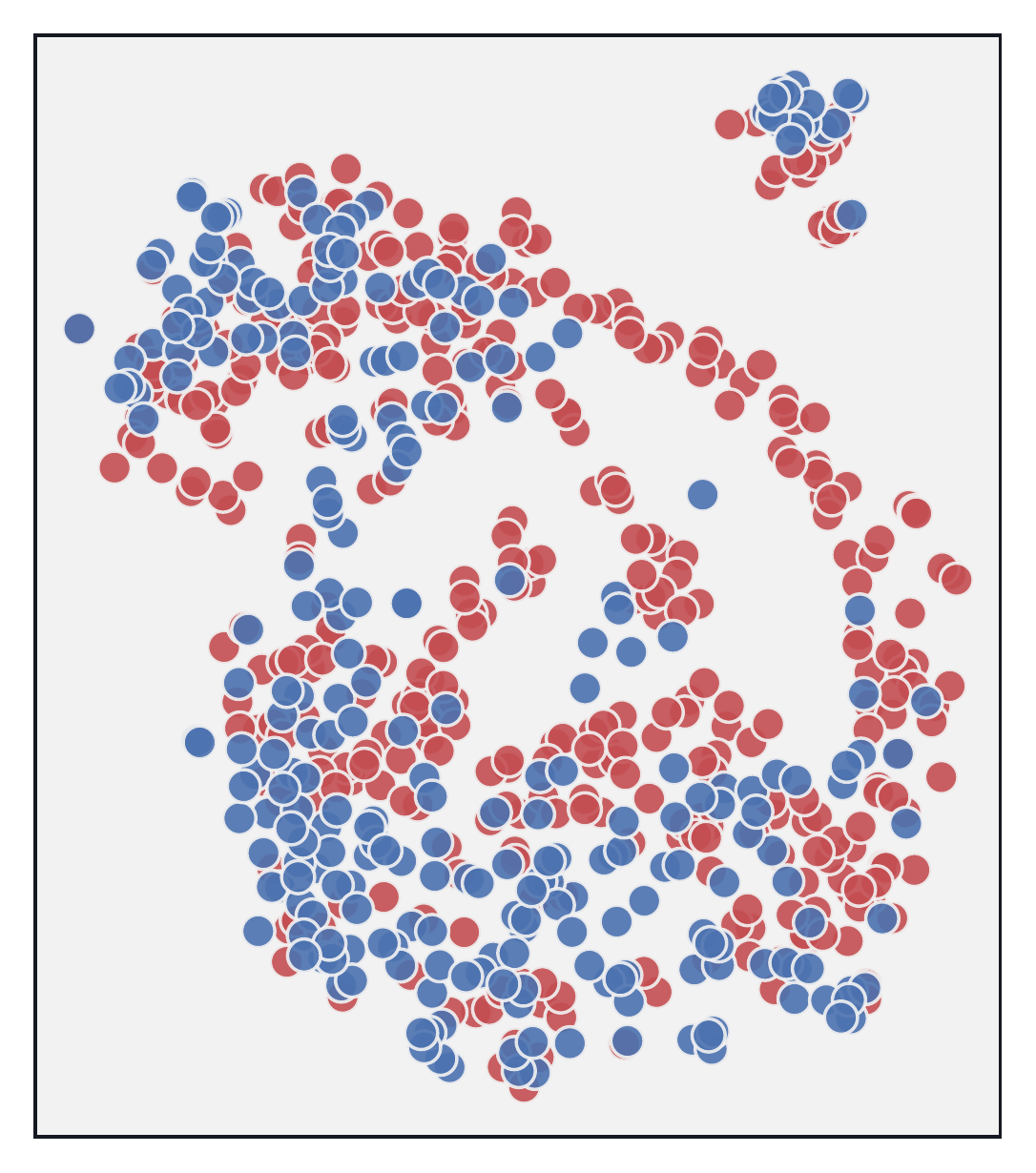}
\includegraphics[width=0.195\textwidth]{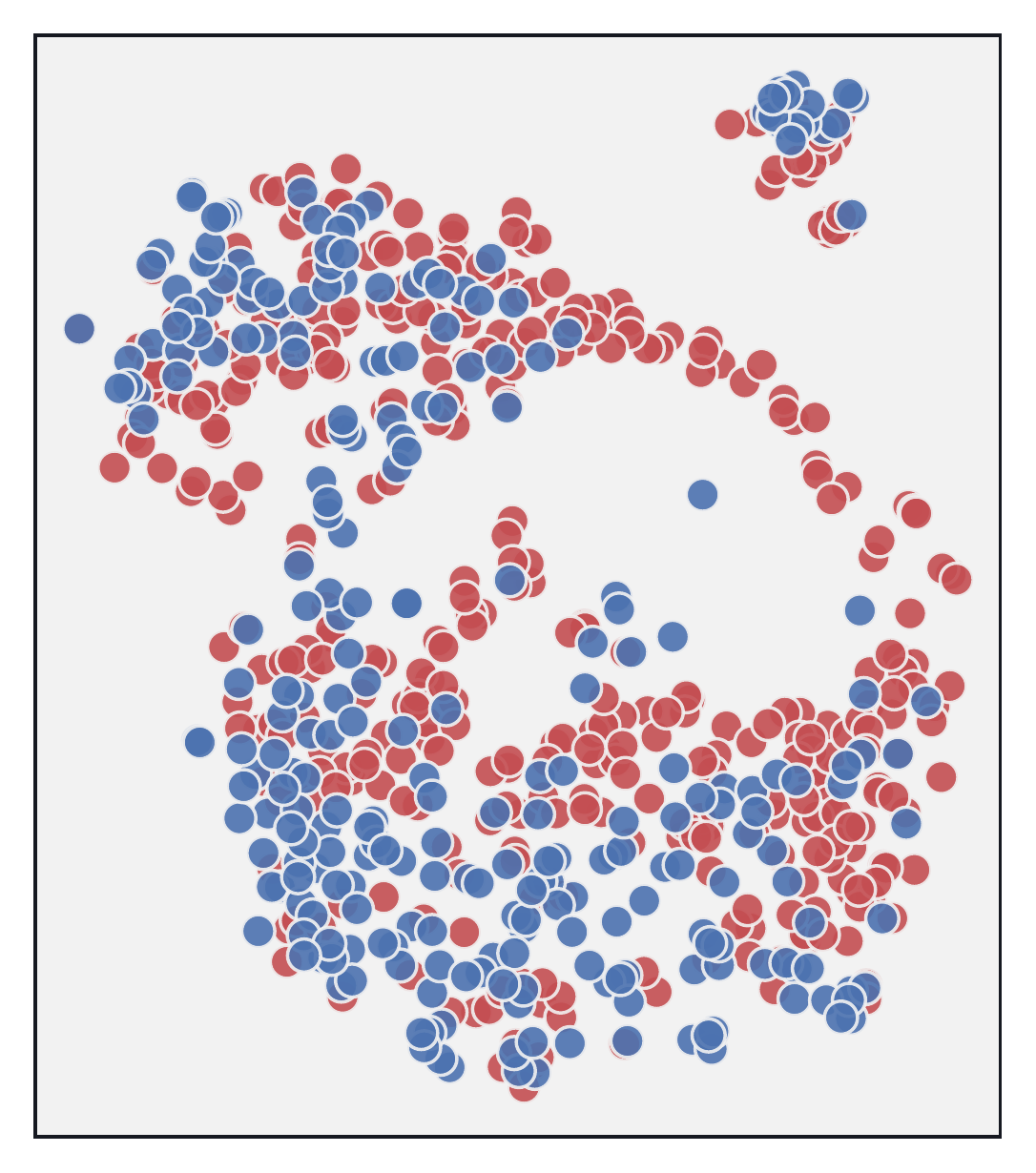}
\caption{Visualized impact of the $\gamma$ parameter of RBU on the shape of the mutual class potential and resulting undersampling regions. Top row: original dataset with highlighted majority objects selected for undersampling. Bottom row: dataset after undersampling. Values of $\gamma$, from the left: $1.0$, $2.5$, $5.0$, $10.0$, $25.0$.}
\label{fig:potential}
\end{figure*}

We present the pseudocode of the proposed method in Algorithm~\ref{algorithm:rbu}. In addition to the collection of majority objects $K$ and the collection of minority objects $\kappa$, algorithm has two additional parameters: spread of the individual radial basis function $\gamma$, affecting the range of impact of the associated observation on the mutual class potential, and the undersampling ratio, with radio equal to $1.0$ indicating that the majority objects are undersampled up to the point of achieving balanced class distribution. Furthermore, we present a visualization of the algorithms behavior for different values of $\gamma$ in Figure~\ref{fig:potential}. As can be seen, the value of $\gamma$ parameter significantly impacts the shape of the resulting potential: using smaller values of $\gamma$ leads to a more complex potential field, affected in a given point in space only by the observations in its close proximity, whereas using larger values of $\gamma$ leads to a smooth potential. As a result, the choice of $\gamma$ affects the order of undersampling. For the smaller values of $\gamma$ removed observations are mostly a part of local clusters consisting of several majority and no minority observations, and these clusters are never completely removed. Furthermore, individual majority observations and majority observations located in a close proximity of minority observations tend to remain unaffected. On the other hand, for larger values of $\gamma$ a single cluster with a high concentration of majority objects is identified and the undersampling is performed solely within its bounds. When combined with a significant data imbalance this can lead to a potentially undesirable behavior of a very highly centralized undersampling, which may indicate the proclivity towards using lower values of $\gamma$.

\begin{algorithm}
\caption{Radial-Based Undersampling}
\label{algorithm:rbu}
\begin{algorithmic}[1]
\STATE \textbf{Input:} collections of majority objects $K$ and minority objects $\kappa$
\STATE \textbf{Parameters:} spread of radial basis function $\gamma$, oversampling $ratio$
\STATE \textbf{Output:} undersampled collection of majority objects $K'$
\STATE
\STATE \textbf{function} RBU($K$, $\kappa$, $\gamma$, $ratio$):
\STATE $K' \gets K$
\FOR{every majority object $K_i'$ in $K'$ and its associated potential $\Phi_i$}
\STATE $\Phi_i \gets \Phi(K_i', K, \kappa, \gamma)$
\ENDFOR
\WHILE{$\vert K \vert - \vert K' \vert < ratio \cdot (\vert K \vert - \vert \kappa \vert)$}
\STATE $x \gets $ majority object $K_i'$ from $K'$ with highest potential $\Phi_i$; in case of multiple selected objects break ties arbitrarily
\STATE discard $x$ from $K'$
\FOR{every majority object $K_i'$ in $K'$ and its associated potential $\Phi_i$}
\STATE $\Phi_i \gets \Phi_i - e^{-(\frac{\lVert K_i' - x \rVert_2}{\gamma})^{2}}$
\ENDFOR
\ENDWHILE
\STATE \textbf{return} $K'$
\end{algorithmic}
\end{algorithm}

\subsection{Computational complexity analysis}

Let us define the total number of observations by $n$, the number of majority and minority observations by $n_{maj}$ and $n_{min}$, respectively, and the number of features by $m$. Let us consider the computational complexity of RBU algorithm applied up to the point of achieving a balanced class distribution. A single calculation of the mutual class potential in any given point, as defined in Equation~\ref{eq:potential}, requires $n$ distance calculations, each with a complexity of $\mathcal{O}(m)$, $n$ summations and $n$ radial basis function calculations, both with a complexity of $\mathcal{O}(1)$. Therefore, a total complexity of a single mutual class potential calculation is equal to $\mathcal{O}(mn)$. Furthermore, the procedure of removing a single observation consists of finding the observation with a highest potential in a collection of a size not exceeding $n_{maj}$, with a complexity equal to $\mathcal{O}(n_{maj})$, discarding it from said collection, the complexity of which we will assume to be $\mathcal{O}(n_{maj})$, and updating all of the remaining potentials, consisting of $n_{maj}$ distance calculations, subtractions and radial basis function calculations, leading to an overall complexity of the potential update operation equal to $\mathcal{O}(mn_{maj})$. Combining the above, the complexity of the procedure of removing a single observation is also equal to $\mathcal{O}(mn_{maj})$. The complete RBU algorithm consists of the initial calculation of the potential for every majority observation, with a complexity of $\mathcal{O}(mnn_{maj})$, and a removal of $n_{maj} - n_{min}$ observations, with a complexity of $\mathcal{O}(mn_{maj}(n_{maj} - n_{min}))$. As a results, the total complexity of the proposed RBU algorithm can be simplified to $\mathcal{O}(mn^2)$. For comparison, the complexity of the original Radial-Based Oversampling algorithm applied up to the point of achieving balanced class distributions, as discussed in \cite{koziarski2019mcrbo}, is equal to $\mathcal{O}(imn^2)$, with $i$ denoting the number of algorithms iterations, the value of which was experimentally chosen to be in range from 1000 to 8000 in the conducted experiments. As a result, RBU has a significantly reduced computational overhead when compared to the original RBO algorithm.

\section{Experimental Study}
\label{sec:exp}

To empirically evaluate the applicability of the proposed Radial-Based Undersampling algorithm we conducted a two-part experimental study. In its first stage we examined the impact of the algorithms parameters on its performance. In the second stage we compared the algorithm with the selected state-of-the-art resampling strategies. Finally, we analysed the parameters of the datasets on which the proposed method achieved the best results to identify possible areas of applicability. In the remainder of this section we describe our experimental set-up and present the observed results.

\subsection{Set-up}
\label{sec:set-up}

\textbf{Data.} Conducted experimental study was based on the binary imbalanced datasets provided in the KEEL repository \cite{alcala2011keel}. Specifically, from the available datasets we excluded the ones containing less than 12 minority observations to avoid issues with cross-validation, as well as the ones for which AUC greater than 0.85 was achieved with SVM without any resampling, to eliminate the datasets for which, despite data imbalance, resampling was not required. A total of 50 datasets was selected using this approach. Out of them, 20 were randomly chosen for the preliminary analysis, during which the impact of the parameters on the algorithms performance was examined. Remaining 30 datasets were used during the comparison with the reference methods.

The details of the used datasets were presented in Table~\ref{table:datasets}. In addition to the imbalance ratio (IR), the number of samples and the number of features, for each dataset we computed the proportion of different types of minority class observations, proposed by Napierała and Stefanowski \cite{napierala2016types}. Specifically, the types were identified using 5-neighbourhood computed based on the Minkowski metric.

Prior to resampling and classification, categorical features were encoded as integers. Afterwards, all features were standarized by removing the mean and scaling to unit variance. No further preprocessing was applied.

\textbf{Classification.} Four different classification algorithms, representing different learning paradigms, were used throughout the experimental study. Specifically, we used CART decision tree, k-nearest neighbors classifier (KNN), naive Bayes classifier (NB) and support vector machine with RBF kernel (SVM). The implementations of the classification algorithms provided in the scikit-learn machine learning library \cite{pedregosa2011scikit} were used, and their default parameters remained unchanged.

\textbf{Reference resampling methods.} During the comparison with reference resampling algorithms we considered a total of 17 different data-level approaches. We focused on other undersampling methods, with a total of 11 algorithms of that type. Specifically, we used random undersampling (RUS), All k-Nearest Neighbors editing (AKNN) \cite{tomek1976experiment}, Cluster Centroid undersampling (CC) \cite{yen2009cluster}, Condensed Nearest Neighbour editing (CNN) \cite{hart1968condensed}, Edited Nearest Neighbour rule (ENN) \cite{wilson1972asymptotic}, Instance Hardness Threshold method (IHT) \cite{smith2014instance}, Neighborhood Cleaning Rule (NCL) \cite{laurikkala2001improving}, Near Miss method (NM) \cite{mani2003knn}, One-Sided Selection (OSS) \cite{Kubat:1997}, Repeated Edited Nearest Neighbour method (RENN) \cite{tomek1976experiment} and Tomek links undersampling (TL) \cite{tomek1976two}. Furthermore, we used 4 additional oversampling algorithms: random oversampling (ROS), SMOTE \cite{chawla2002smote}, Borderline-SMOTE (Bord) \cite{han2005borderline} and Radial-Based Oversampling (RBO) \cite{koziarski2019mcrbo}. Finally, we also used two methods combining oversampling with undersampling: SMOTE combined with Tomek links (STL) \cite{tomek1976two} and Edited Nearest Neighbor rule (SENN) \cite{wilson1972asymptotic}.

With the exception of RBO, the implementations of the reference methods provided in the imbalanced-learn library \cite{lemaitre2017imbalanced} were used.

\textbf{Evaluation.} For every dataset we reported the results averaged over the $5\times2$ cross-validation folds \cite{alpaydin1999combined}. Throughout the experimental study we reported the values of precision, recall, F-measure, AUC and G-mean. Whenever applicable, parameter selection was conducted by further $3\times2$ cross-validation on the currently considered training data, with the optimization criterion being the average of F-measure, AUC and G-mean. It should be noted that in most cases observed F-measure was significantly lower than AUC and G-mean, leading to the choice of parameters biased towards the latter two metrics.

\textbf{Implementation and reproducibility.} The experiments described in this paper were implemented in the Python programming language. Complete code, sufficient to repeat the experiments, was made publicly available at\footnote{\url{https://github.com/michalkoziarski/RBU}}. In addition to the code we also provided the cross-validation folds used during the experiments, as well as a file containing complete results, enabling any further analysis.

\begin{table*}
\small
\caption{Details of the datasets used during the preliminary (top) and the final (bottom) analysis.}
\label{table:datasets}
\centering
\setlength{\tabcolsep}{5pt}
\begin{tabular}{llllllll}
\toprule
\textbf{Name} & \textbf{IR} & \textbf{Samples} & \textbf{Features} & \textbf{Safe [\%]} & \textbf{Borderline [\%]} & \textbf{Rare [\%]} & \textbf{Outlier [\%]} \\
\midrule
pima & 1.87 & 768 & 8 & 30.22 & 44.03 & 16.79 & 8.96 \\
glass0 & 2.06 & 214 & 9 & 54.29 & 38.57 & 1.43 & 5.71 \\
vehicle3 & 2.99 & 846 & 18 & 16.51 & 50.94 & 27.36 & 5.19 \\
ecoli1 & 3.36 & 336 & 7 & 53.25 & 31.17 & 9.09 & 6.49 \\
yeast3 & 8.1 & 1484 & 8 & 56.44 & 25.15 & 7.36 & 11.04 \\
ecoli-0-6-7\_vs\_3-5 & 9.09 & 222 & 7 & 40.91 & 31.82 & 9.09 & 18.18 \\
yeast-0-3-5-9\_vs\_7-8 & 9.12 & 506 & 8 & 16.0 & 28.0 & 22.0 & 34.0 \\
ecoli-0-2-6-7\_vs\_3-5 & 9.18 & 224 & 7 & 40.91 & 31.82 & 9.09 & 18.18 \\
ecoli-0-1-4-7\_vs\_2-3-5-6 & 10.59 & 336 & 7 & 65.52 & 17.24 & 0.0 & 17.24 \\
glass-0-1-4-6\_vs\_2 & 11.06 & 205 & 9 & 0.0 & 17.65 & 35.29 & 47.06 \\
cleveland-0\_vs\_4 & 12.31 & 173 & 13 & 0.0 & 69.23 & 23.08 & 7.69 \\
yeast-2\_vs\_8 & 23.1 & 482 & 8 & 55.0 & 0.0 & 15.0 & 30.0 \\
winequality-red-4 & 29.17 & 1599 & 11 & 0.0 & 7.55 & 22.64 & 69.81 \\
winequality-red-8\_vs\_6 & 35.44 & 656 & 11 & 0.0 & 0.0 & 50.0 & 50.0 \\
kr-vs-k-zero\_vs\_eight & 53.07 & 1460 & 6 & 62.96 & 25.93 & 7.41 & 3.7 \\
winequality-white-3-9\_vs\_5 & 58.28 & 1482 & 11 & 0.0 & 8.0 & 20.0 & 72.0 \\
poker-8-9\_vs\_6 & 58.4 & 1485 & 10 & 8.0 & 56.0 & 20.0 & 16.0 \\
abalone-20\_vs\_8-9-10 & 72.69 & 1916 & 8 & 0.0 & 19.23 & 11.54 & 69.23 \\
poker-8\_vs\_6 & 85.88 & 1477 & 10 & 5.88 & 35.29 & 35.29 & 23.53 \\
abalone19 & 129.44 & 4174 & 8 & 0.0 & 0.0 & 12.5 & 87.5 \\
\midrule
glass1 & 1.82 & 214 & 9 & 44.74 & 32.89 & 14.47 & 7.89 \\
yeast1 & 2.46 & 1484 & 8 & 21.91 & 45.69 & 20.75 & 11.66 \\
haberman & 2.78 & 306 & 3 & 4.94 & 48.15 & 32.1 & 14.81 \\
vehicle1 & 2.9 & 846 & 18 & 23.96 & 55.76 & 16.13 & 4.15 \\
ecoli3 & 8.6 & 336 & 7 & 28.57 & 48.57 & 8.57 & 14.29 \\
yeast-2\_vs\_4 & 9.08 & 514 & 8 & 54.9 & 19.61 & 11.76 & 13.73 \\
yeast-0-2-5-6\_vs\_3-7-8-9 & 9.14 & 1004 & 8 & 33.33 & 32.32 & 14.14 & 20.2 \\
yeast-0-5-6-7-9\_vs\_4 & 9.35 & 528 & 8 & 7.84 & 43.14 & 15.69 & 33.33 \\
ecoli-0-6-7\_vs\_5 & 10.0 & 220 & 6 & 45.0 & 35.0 & 0.0 & 20.0 \\
glass-0-1-6\_vs\_2 & 10.29 & 192 & 9 & 0.0 & 29.41 & 35.29 & 35.29 \\
glass2 & 11.59 & 214 & 9 & 0.0 & 23.53 & 41.18 & 35.29 \\
yeast-1\_vs\_7 & 14.3 & 459 & 7 & 6.67 & 33.33 & 26.67 & 33.33 \\
glass4 & 15.46 & 214 & 9 & 23.08 & 53.85 & 7.69 & 15.38 \\
page-blocks-1-3\_vs\_4 & 15.86 & 472 & 10 & 78.57 & 17.86 & 3.57 & 0.0 \\
abalone9-18 & 16.4 & 731 & 8 & 4.76 & 23.81 & 19.05 & 52.38 \\
yeast-1-4-5-8\_vs\_7 & 22.1 & 693 & 8 & 0.0 & 6.67 & 33.33 & 60.0 \\
flare-F & 23.79 & 1066 & 11 & 0.0 & 48.84 & 39.53 & 11.63 \\
car-good & 24.04 & 1728 & 6 & 0.0 & 97.1 & 2.9 & 0.0 \\
car-vgood & 25.58 & 1728 & 6 & 20.0 & 80.0 & 0.0 & 0.0 \\
yeast4 & 28.1 & 1484 & 8 & 7.84 & 35.29 & 17.65 & 39.22 \\
yeast-1-2-8-9\_vs\_7 & 30.57 & 947 & 8 & 3.33 & 23.33 & 23.33 & 50.0 \\
yeast5 & 32.73 & 1484 & 8 & 34.09 & 50.0 & 11.36 & 4.55 \\
abalone-17\_vs\_7-8-9-10 & 39.31 & 2338 & 8 & 3.45 & 13.79 & 34.48 & 48.28 \\
abalone-21\_vs\_8 & 40.5 & 581 & 8 & 14.29 & 35.71 & 21.43 & 28.57 \\
yeast6 & 41.4 & 1484 & 8 & 34.29 & 25.71 & 11.43 & 28.57 \\
winequality-white-3\_vs\_7 & 44.0 & 900 & 11 & 0.0 & 15.0 & 10.0 & 75.0 \\
abalone-19\_vs\_10-11-12-13 & 49.69 & 1622 & 8 & 0.0 & 0.0 & 21.88 & 78.12 \\
kddcup-buffer\_overflow\_vs\_back & 73.43 & 2233 & 41 & 73.33 & 13.33 & 6.67 & 6.67 \\
poker-8-9\_vs\_5 & 82.0 & 2075 & 10 & 0.0 & 0.0 & 16.0 & 84.0 \\
kddcup-rootkit-imap\_vs\_back & 100.14 & 2225 & 41 & 54.55 & 27.27 & 9.09 & 9.09 \\
\bottomrule
\end{tabular}
\end{table*}

\subsection{Analysis of the impact of parameters}

In the first stage of the conducted experimental study we considered the impact of two of the proposed algorithms parameters, radial basis function spread $\gamma$ and the undersampling ratio, on the algorithms performance. Specifically, we conducted two experiments: in the first one we adjusted the value of $\gamma$ parameter in \{0.001, 0.01, 0.1, 1.0, 10.0, 100.0\} while selecting the values of undersampling ratio individually for each dataset using cross-validation, with considered values in \{0.0, 0.2, 0.4, 0.6, 0.8, 1.0\}. In the second experiment we used the same parameter values, but adjusted undersampling ratio while selecting $\gamma$ with cross-validation.

\begin{figure*}
\centering
\includegraphics[width=0.49\textwidth]{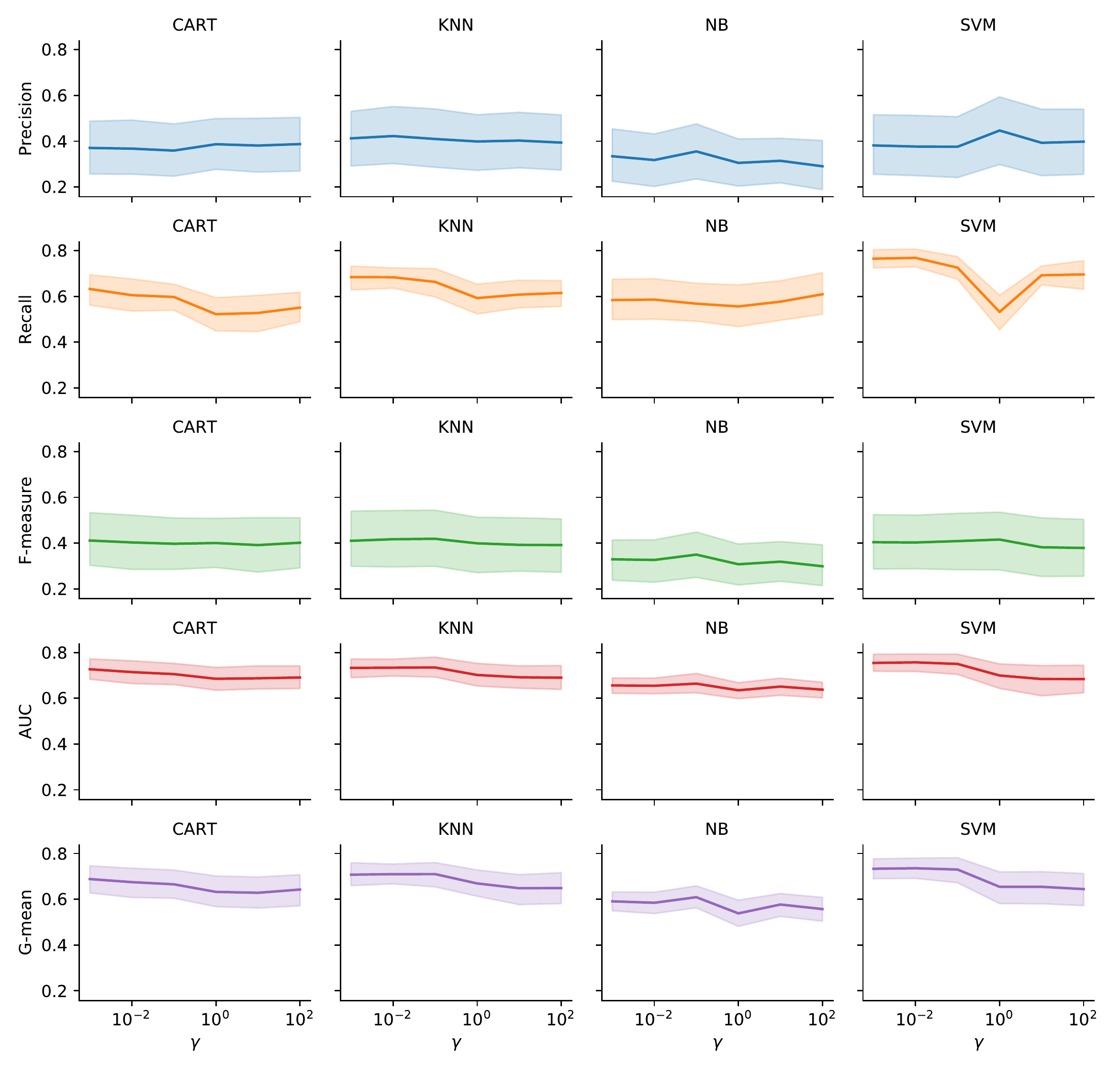}
\includegraphics[width=0.49\textwidth]{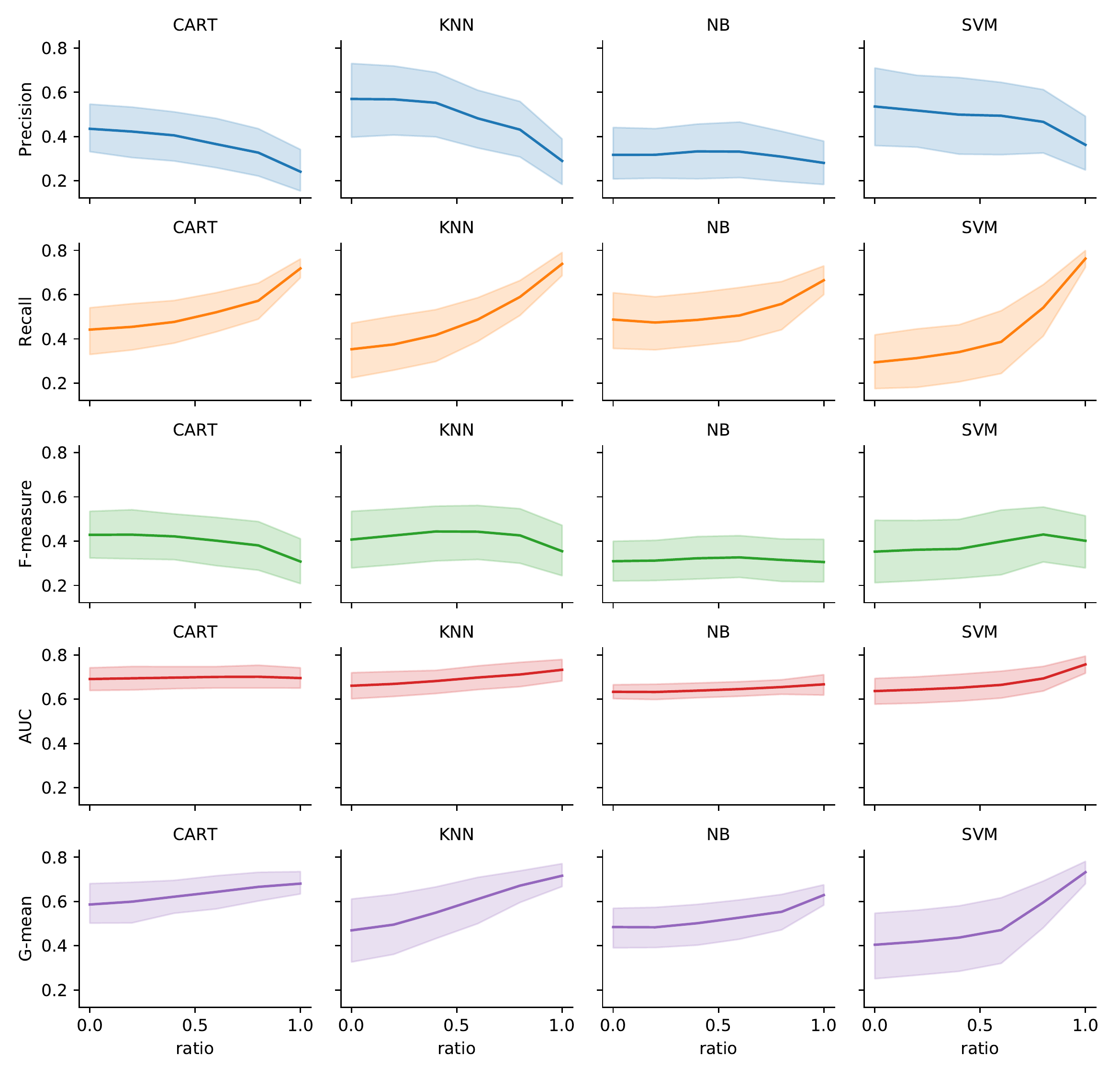}
\caption{The impact of $\gamma$ parameter (left) and undersampling ratio (right) of Radial-Based Undersampling on various performance metrics, averaged over all datasets, with a 95\% confidence intervals shown.}
\label{fig:prelim}
\end{figure*}

The results, averaged over 20 datasets, were presented in Figure~\ref{fig:prelim}. As can be seen, the best performance with respect to the combined metrics (F-measure, AUC, G-mean) was observed for smaller values of $\gamma$, equal to 0.1 or lower. Using higher values either did not improve the average performance, or resulted in its significant decline. The exact value of $\gamma$ parameter for which the best averaged performance was observed depended on the type of classifier and the chosen metric. For CART, KNN and SVM classifiers decreasing the value of $\gamma$ tended to improve the recall at the cost of precision, whereas for NB the reverse was observed. It is worth noting that while the described trends were roughly monotonic for individual classifier and metric combinations, a significant peak in precision, combined with a drop in recall, was observed in the case of SVM for $\gamma = 1.0$. During the examination of the results for individual datasets it was confirmed that this peak occurred in about half of the datasets. It is not clear what caused the peak and what is its significance.

In the case of the undersampling ratio, as can be expected, increasing the ratio led to an improvement in the classifiers recall and a corresponding drop in the precision, for all of the examined classification algorithms. When considering the combined metrics, the relation between performance of the algorithm and the undersampling ratio varied depending on the metric. In the case of the G-mean, a significantly better performance was observed for the highest value of the undersampling ratio, corresponding to undersampling up to the point of achieving balanced class distribution. A noticeable improvement in performance was also observed in the case of AUC in combination with KNN ans SVM classifiers. When combined with CART and NB classifiers, the undersampling ratio did not, on average, have a significant impact on the observed AUC. In the case of F-measure the peak in performance was observed for different undersampling ratios. In the case of SVM high, but not complete, undersampling led to achieving the best results, with the best performance observed for the undersampling ratio of 0.8. In the case of KNN and NB best results were achieved for medium ratios of 0.4 and 0.6, but in the case of NB the impact of the choice of the undersampling ratio was less significant. Finally, in the case of CART the best performance was observed for small or no undersampling, with ratios equal to 0.0 and 0.2. Notably, in no case was the best performance with regard to F-measure observed for complete undersampling, up to the point of achieving balanced class distributions.

To summarize, irregardless of the choice of the classification algorithm and the evaluation metric, the best performance was observed for smaller values of $\gamma$ parameter in \{0.001, 0.01, 0.1\}, which could all be use as a sensible default values. The choice of the undersampling ratio, however, was dependant on the classification algorithm and evaluation metric. While using complete undersampling was a sensible default with regard to AUC and G-mean, lower undersampling ratios led to observing better performance with regard to F-measure, and the choice of the classifier affected the exact value of the undersampling ratio for which the best performance was observed. It is worth noting that the optimization of parameters with respect to one of the metrics could lead to suboptimal results with respect to the other. In particular, when using the scheme employed in this paper, that is choosing the parameters maximizing the average value of F-measure, AUC and G-mean, the parameter choice will be biased towards the AUC and G-mean: this is both because the F-measure tends to take lower values, and the fact that the AUC and G-mean displayed higher correlation between each other than between the F-measure.

\subsection{Comparison with other methods}

In the second stage of the conducted experimental analysis we compared the proposed Radial-Based Undersampling with a total of 17 data-level methods described in Section~\ref{sec:set-up}. For every algorithm we conducted a parameter search to adjust its hyperparameters individually for each dataset. Specifically, for all variants of SMOTE we considered the values of $k$ neighborhood in \{1, 3, 5, 7, 9\}; for Bord, we additionally considered the values of $m$ neighborhood, used to determine if a minority sample is in danger, in \{5, 10, 15\}; for neighborhood-based undersampling strategies, that is AKNN, CNN, ENN, NCL, NM, OSS and RENN we considered the values of respective $k$ neighborhoods in \{1, 3, 5, 7\}; for RBU and RBO we considered the values of $\gamma$ in \{0.01, 0.1, 1.0, 10.0\}. Finally, for all of the algorithms using manually specified resampling ratio, that is RUS, ROS, CC, RBO, RBU and all of the variants of SMOTE, we considered the values of that ratio in \{0.5, 0.75, 1.0\}. To evaluate the statistical significance of the observed results we used the Friedman test combined with the Shaffer's post-hoc. The results were reported at the significance level $\alpha = 0.10$.

\begin{figure}
\centering
\includegraphics[width=0.95\linewidth]{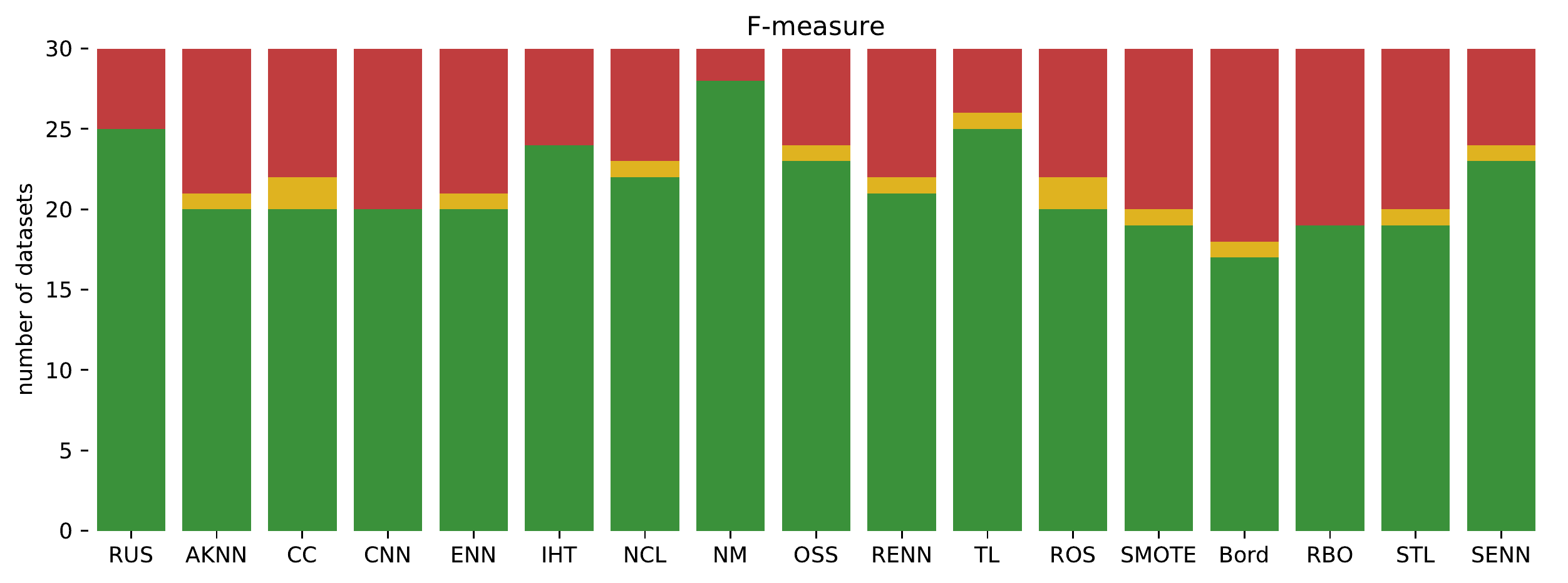}
\includegraphics[width=0.95\linewidth]{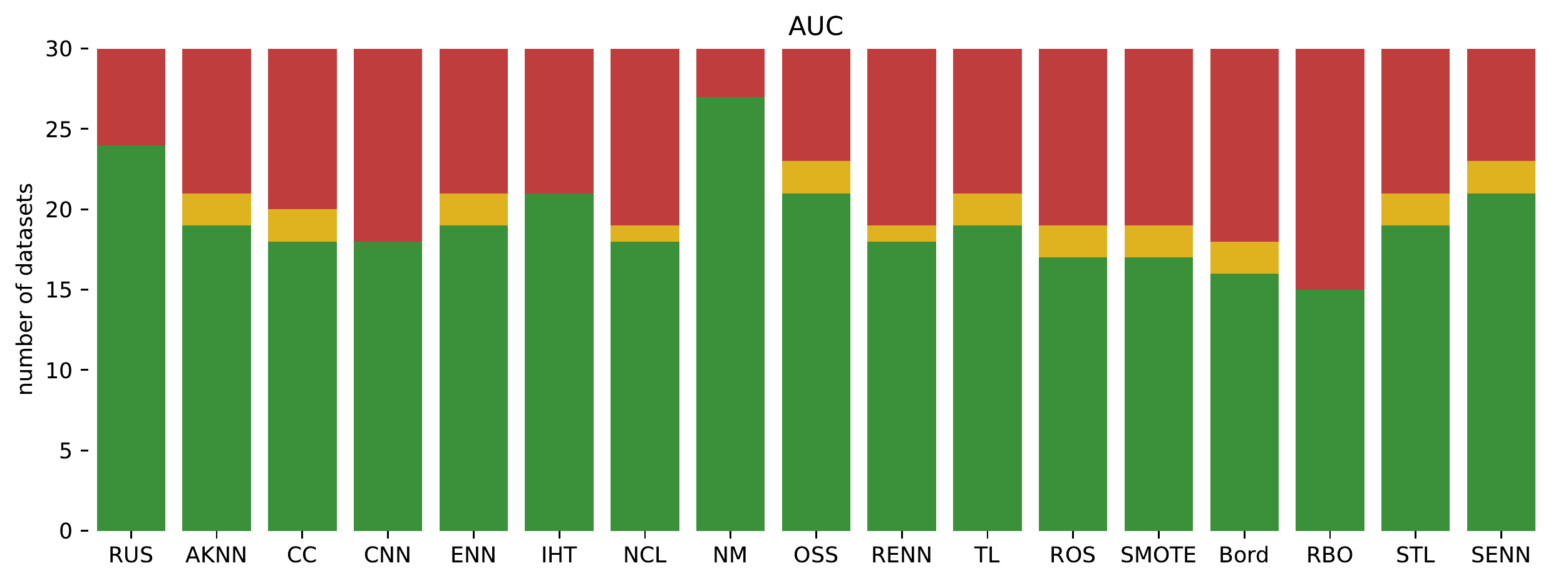}
\includegraphics[width=0.95\linewidth]{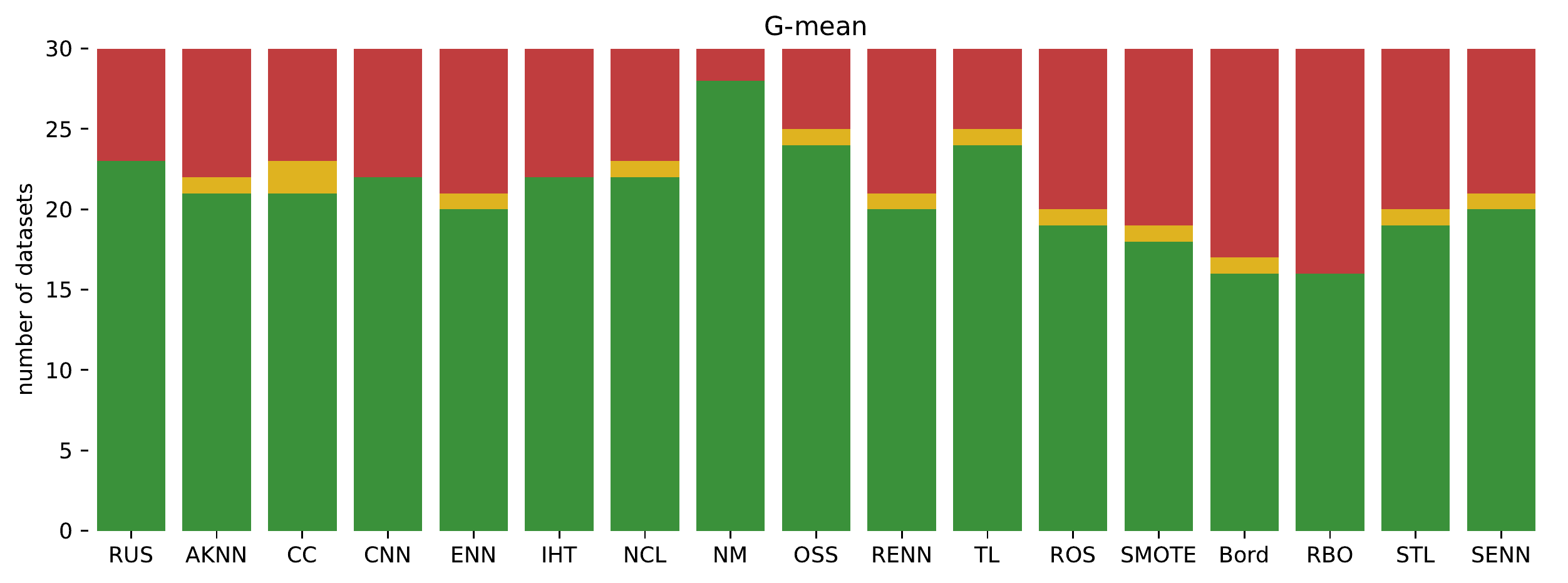}
\caption{Total number of datasets for which RBU achieved better (green), equal (yellow) or worse (red) performance than specific reference methods, with naive Bayes algorithm used for classification.}
\label{fig:results-win-loss-tie}
\end{figure}

We present the average rankings achieved by the respective methods for all of the performance metrics and highlight the statistically significantly different results in Table~\ref{table:results-final}. Furthermore, for NB classifier we present a detailed win-tie-loss analysis, that is a visualization of the number of datasets on which RBU outperforms individual reference methods, in Figure~\ref{fig:results-win-loss-tie}. As can be seen, in general case the usefulness of the proposed RBU algorithm, when compared to the reference methods, was reliant on both the choice of the classification algorithm and the metric used to evaluate the performance. RBU achieved the best results when combined with NB classifier, scoring highest average ranks with respect to all of the combined performance metrics, and statistically significantly better results with respect to at least one of them for 10 out of 17 reference methods. Furthermore, it achieved a relatively good performance when combined with CART and SVM, scoring a statistically significantly better results with respect to at least one of the combined metrics: in 6 cases for CART, and in 7 cases for SVM. At the same time, for both classifiers a statistically significantly worse results were observed in only a single case: compared to ENN for CART classifier, and compared to SMOTE for SVM, both with respect to F-measure. The worse performance was observed when RBU was combined with KNN classifier, in which case all of the variants of SMOTE, as well as the NCL algorithm, achieved a statistically significantly better results than RBU. In that case, the latter significantly outperformed only two of the reference methods.

When compared to the RBO, RBU achieved a statistically significantly different results only in case of CART decision tree. In that instance RBU achieved a significantly better AUC and G-mean than RBO. Out of the remaining cases the highest disproportion in average ranks was observed in combination with SVM, this time in favor of RBO, but the results were not statistically significant.

It is worth noting that for CART, KNN and SVM classifiers RBU achieved a higher rank with respect to recall than the rank achieved with respect to precision, whereas the opposite was true for NB. This indicates that undersampling with RBU affects NB classification differently than the remaining classifiers, and has less severe impact on the precision of that classifier.

To summarize, while the proposed RBU algorithm, in general case, did not achieve the best results when applied in combination with all of the considered classification algorithms, it performed best when combined with NB classifier, and to a lesser extent with CART and SVM. The areas of applicability with respect to the choice of the classification algorithm partially overlap for RBU and RBO: both algorithms displayed comparatively good performance when combined with the NB classifier, but RBU scored significantly better results when combined with CART. Finally, for CART, KNN and SVM classifiers RBU achieved comparatively better recall than precision, but the opposite was true for the NB classifier.

\begin{sidewaystable*}
\tiny
\caption{Average rankings of the evaluated methods. Best performance was denoted with bold font. Methods that achieved significantly different results (according to Shaffer's post-hoc test) than RBU where denoted in subscript: with $+$ sign for methods compared to which RBU achieved a better results, and $-$ sign for methods compared to which RBU achieved worse results.}
\label{table:results-final}
\centering
\begin{tabularx}{\textwidth}{llllllllllllllllllll}
\toprule
& & & \multicolumn{11}{l}{\textbf{Undersampling}} & \multicolumn{4}{l}{\textbf{Oversampling}} & \multicolumn{2}{l}{\textbf{Combined}} \\
\midrule
& Metric & RBU & RUS & AKNN & CC & CNN & ENN & IHT & NCL & NM & OSS & RENN & TL & ROS & SMOTE & Bord & RBO & STL & SENN \\
\midrule
\parbox[t]{2mm}{\multirow{5}{*}{\rotatebox[origin=c]{90}{CART}}} & Precision & 11.4 & 15.2 & 8.8 & 15.7 & 14.7 & 8.0 & 14.3 & 8.2 & 17.2 \textsubscript{+} & 4.9 \textsubscript{--} & 9.8 & \first{3.9} \textsubscript{--} & \second{4.4} \textsubscript{--} & 6.8 \textsubscript{--} & 6.4 \textsubscript{--} & 5.1 \textsubscript{--} & 6.8 \textsubscript{--} & 9.4 \\
 & Recall & 7.9 & 4.6 & 8.0 & \second{3.8} & 9.2 & 9.0 & \first{3.4} \textsubscript{--} & 9.3 & 4.7 & 13.4 \textsubscript{+} & 7.6 & 13.0 \textsubscript{+} & 15.8 \textsubscript{+} & 12.9 \textsubscript{+} & 14.1 \textsubscript{+} & 14.5 \textsubscript{+} & 12.1 & 7.7 \\
 & F-measure & 10.1 & 13.5 & 6.3 & 14.2 & 14.5 & \first{5.5} \textsubscript{--} & 11.8 & 5.9 & 16.7 \textsubscript{+} & 6.5 & 7.5 & \second{5.6} & 9.0 & 9.0 & 9.2 & 9.1 & 8.9 & 7.6 \\
 & AUC & 7.6 & 6.7 & 7.1 & 8.7 & 12.7 \textsubscript{+} & 7.5 & \first{6.0} & 7.5 & 14.0 \textsubscript{+} & 10.8 & 6.3 & 10.2 & 13.6 \textsubscript{+} & 11.0 & 12.0 & 12.8 \textsubscript{+} & 10.2 & \second{6.2} \\
 & G-mean & 7.2 & \first{5.0} & 7.5 & 7.2 & 10.7 & 8.2 & \second{5.9} & 8.3 & 11.3 & 12.2 \textsubscript{+} & 7.2 & 11.6 & 14.4 \textsubscript{+} & 11.3 & 12.5 \textsubscript{+} & 13.2 \textsubscript{+} & 10.8 & 6.3 \\
\midrule
\parbox[t]{2mm}{\multirow{5}{*}{\rotatebox[origin=c]{90}{KNN}}} & Precision & 11.9 & 14.2 & 7.0 \textsubscript{--} & 13.5 & 8.2 & 6.6 \textsubscript{--} & 10.8 & 6.6 \textsubscript{--} & 14.2 & \second{4.7} \textsubscript{--} & 8.2 & \first{4.6} \textsubscript{--} & 8.6 & 9.6 & 9.2 & 8.7 & 10.8 & 13.7 \\
 & Recall & 8.8 & 4.9 & 12.5 & \second{4.7} & 12.5 & 13.7 \textsubscript{+} & 7.8 & 12.4 & 8.3 & 17.0 \textsubscript{+} & 11.8 & 17.3 \textsubscript{+} & 8.9 & 6.1 & 7.6 & 8.6 & 5.1 & \first{3.1} \textsubscript{--} \\
 & F-measure & 11.4 & 11.8 & 9.0 & 11.1 & 9.6 & 9.4 & 9.4 & 6.7 \textsubscript{--} & 12.9 & 13.0 & 9.8 & 13.3 & 7.0 & \second{6.5} \textsubscript{--} & \first{6.1} \textsubscript{--} & 7.0 & 7.3 & 9.7 \\
 & AUC & 10.4 & 6.6 & 11.8 & 6.8 & 12.3 & 13.0 & 8.1 & 11.0 & 11.4 & 16.4 \textsubscript{+} & 11.2 & 16.4 \textsubscript{+} & 7.9 & \second{4.6} \textsubscript{--} & 6.6 & 7.4 & \first{4.3} \textsubscript{--} & 4.7 \textsubscript{--} \\
 & G-mean & 9.8 & 6.0 & 12.4 & 6.7 & 12.5 & 13.2 & 7.8 & 11.3 & 10.5 & 16.8 \textsubscript{+} & 11.9 & 17.0 \textsubscript{+} & 7.8 & 4.6 \textsubscript{--} & 6.7 & 7.4 & \first{4.2} \textsubscript{--} & \second{4.5} \textsubscript{--} \\
\midrule
\parbox[t]{2mm}{\multirow{5}{*}{\rotatebox[origin=c]{90}{NB}}} & Precision & \first{6.2} & 11.0 \textsubscript{+} & 7.9 & 13.7 \textsubscript{+} & 8.4 & 8.1 & 10.6 & 10.0 & 15.4 \textsubscript{+} & 8.9 & 7.7 & 8.8 & 10.8 \textsubscript{+} & \second{7.6} & 8.0 & 7.9 & 9.7 & 10.4 \\
 & Recall & 10.8 & 10.4 & 11.3 & \first{4.0} \textsubscript{--} & 11.1 & 11.0 & 8.8 & 10.2 & 11.3 & 11.0 & 11.4 & 11.0 & \second{5.5} \textsubscript{--} & 8.2 & 9.8 & 9.6 & 6.9 & 8.7 \\
 & F-measure & \first{5.6} & 10.8 \textsubscript{+} & 8.4 & 12.0 \textsubscript{+} & 8.8 & 8.6 & 10.3 \textsubscript{+} & 10.1 & 15.6 \textsubscript{+} & 10.3 \textsubscript{+} & 8.8 & 10.3 \textsubscript{+} & 10.6 \textsubscript{+} & 7.3 & \second{7.0} & 7.2 & 8.9 & 10.4 \textsubscript{+} \\
 & AUC & \first{6.6} & 10.0 & 9.0 & 10.5 & 9.3 & 9.2 & 8.5 & 10.4 & 15.6 \textsubscript{+} & 11.6 \textsubscript{+} & 9.5 & 11.0 & 10.4 & 7.8 & 7.7 & \second{6.6} & 8.1 & 9.4 \\
 & G-mean & \first{6.0} & 10.1 & 9.9 & 10.9 \textsubscript{+} & 11.0 \textsubscript{+} & 10.2 & 8.5 & 11.4 \textsubscript{+} & 12.1 \textsubscript{+} & 12.5 \textsubscript{+} & 9.8 & 12.1 \textsubscript{+} & 9.8 & 7.1 & 7.0 & \second{6.1} & 7.4 & 8.8 \\
\midrule
\parbox[t]{2mm}{\multirow{5}{*}{\rotatebox[origin=c]{90}{SVM}}} & Precision & 11.7 & 13.2 & 9.2 & 12.5 & 7.3 & 8.4 & 12.0 & 7.7 & 15.7 & 7.5 & 8.8 & \second{6.7} \textsubscript{--} & 7.9 & 6.8 \textsubscript{--} & \first{6.5} \textsubscript{--} & 7.7 & 9.1 & 12.2 \\
 & Recall & 7.5 & \first{4.1} & 12.0 \textsubscript{+} & 4.2 & 13.1 \textsubscript{+} & 13.5 \textsubscript{+} & 7.1 & 13.2 \textsubscript{+} & 6.0 & 16.6 \textsubscript{+} & 12.3 \textsubscript{+} & 17.1 \textsubscript{+} & 7.7 & 7.8 & 10.0 & 7.7 & 7.0 & \second{4.2} \\
 & F-measure & 9.8 & 11.1 & 9.4 & 9.3 & 11.3 & 9.3 & 9.9 & 9.0 & 14.5 \textsubscript{+} & 15.0 \textsubscript{+} & 9.9 & 14.8 \textsubscript{+} & 6.4 & \first{4.6} \textsubscript{--} & \second{5.5} & 5.7 & 6.8 & 8.9 \\
 & AUC & 8.7 & 5.7 & 11.7 & 5.6 & 12.9 & 12.9 & 8.6 & 12.8 & 12.9 & 16.4 \textsubscript{+} & 11.6 & 16.3 \textsubscript{+} & 5.8 & \second{5.3} & 8.0 & \first{5.2} & 5.7 & \first{5.2} \\
 & G-mean & 7.9 & \second{4.9} & 12.0 & 5.2 & 13.4 \textsubscript{+} & 13.2 \textsubscript{+} & 8.2 & 13.0 \textsubscript{+} & 11.2 & 16.6 \textsubscript{+} & 12.4 \textsubscript{+} & 16.7 \textsubscript{+} & 5.8 & 5.9 & 8.6 & 5.7 & 5.5 & \first{4.7} \\
\bottomrule
\end{tabularx}
\end{sidewaystable*}

\subsection{Analysis of the impact of dataset characteristics}

\begin{figure*}
\centering
\includegraphics[width=0.32\textwidth]{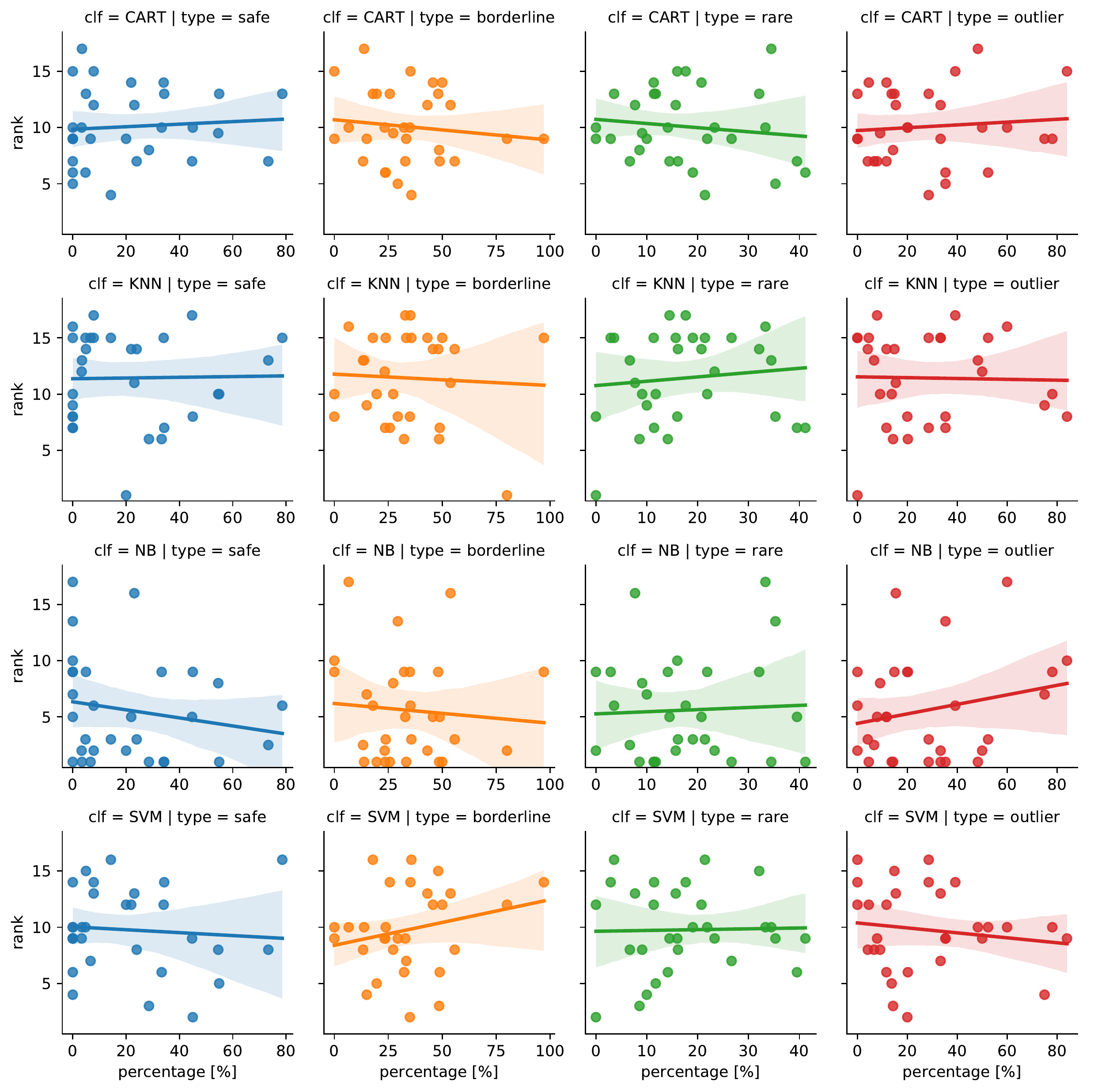}
\includegraphics[width=0.32\textwidth]{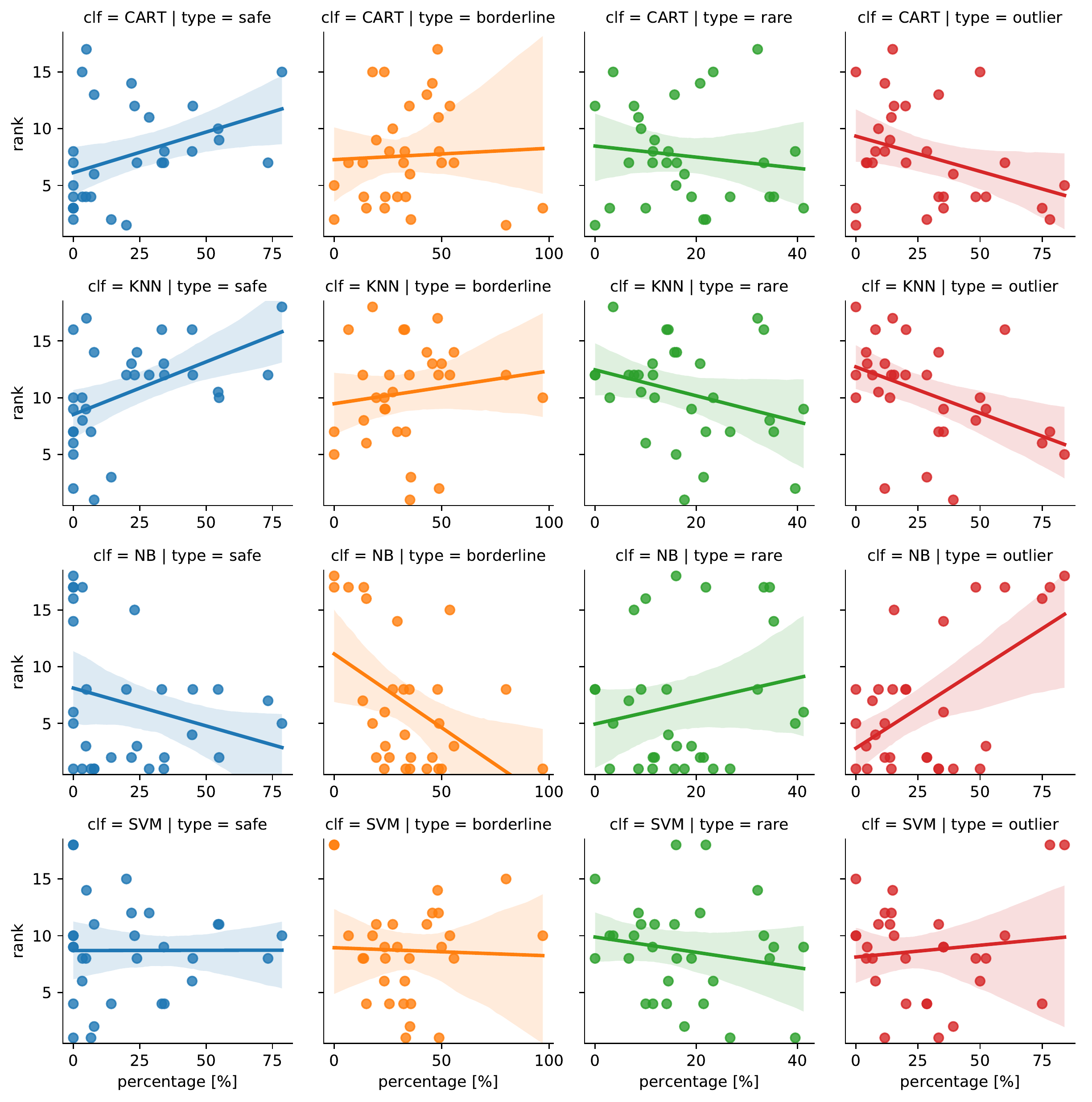}
\includegraphics[width=0.32\textwidth]{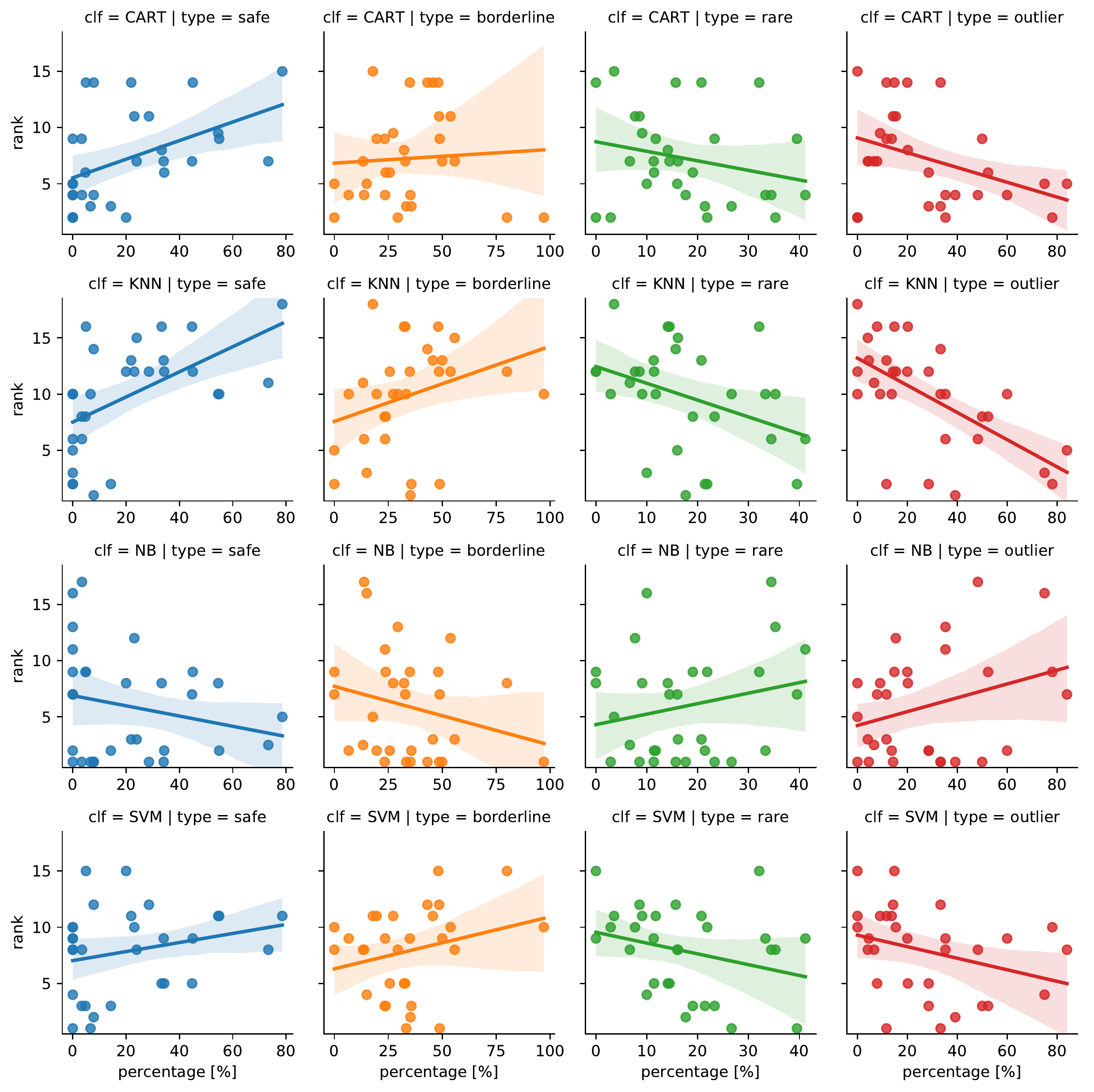}
\caption{Scatterplots representing relation between the percentage of the objects of a given type: safe (blue), borderline (orange), rare (green) and outlier (red), and the rank achieved by RBU on the given dataset, with regard to F-measure (left grid), AUC (middle grid) and G-mean (right grid). Each row contains data for a single classifier, from the top: CART, KNN, NB and SVM. 95\% confidence invervals were shown.}
\label{fig:type}
\end{figure*}

In the final stage of the conducted experimental study we examined if the performance of the algorithm changes depending on the characteristics of the dataset on which it is applied. Specifically, we considered the categorization proposed by Napierała and Stefanowski \cite{napierala2016types} to evaluate the fraction of minority objects belonging to one of the categories: safe, borderline, rare and outlier, for each individual dataset. Afterwards, we examined the relation between the fraction of objects of a given type and the rank the RBU method achieved compared to the reference algorithms. In the Table~\ref{table:type} we present the Pearson correlation coefficient between the percentage of the objects of a given type and the rank obtained for that dataset, with highlighted statistically significant correlations at the significance level $\alpha = 0.10$. Furthermore, in Figure~\ref{fig:type} we present scatterplots containing individual data points with a linear regression model fit. Note that ranking was performed in a descending order, meaning that the best performing method received the rank equal to 1. Therefore, negative correlation and regression slope indicate that the relative rank increases.

Similarly to the results observed during the comparison with reference methods, the trends observed for CART, KNN and SVM classifiers differed from the ones observed for the NB classifier. For all of the three former classifiers a statistically significantly worse recall, compared to the reference methods, was observed when the datasets contained a high proportion of safe objects. Conversely, a statistically significantly better recall was observed for the datasets containing higher proportion of rare and outlier minority objects. This, in turn, led to a significantly higher values of AUC and G-mean for the datasets containing a larger proportion of rare and outlier minority objects, and significantly lower values of these metrics for datasets containing large proportion of safe objects. In the case of NB classifier, on the other hand, a significantly worse results, compared with the reference methods, were observed with regard to AUC and G-mean for datasets with a large proportion of outliers, and significantly better results with regard to AUC for datasets with a large proportion of borderline minority objects. However, no significant relations with regard to precision or recall were observed for NB in those cases.

To summarize, the results of the analysis of the impact of dataset characteristics indicate that the proposed RBU algorithm, when used with CART, KNN or SVM classifier, is particularly well suited for resampling datasets with a high proportion of rare and outlier minority objects, but achieves a relatively worse performance for safe datasets. This is caused mainly due to the differences in the classification recall. However, these trends do not extended to the case of the NB classifier, for which the observed performance with regard to the AUC and G-mean was actually worse when datasets contained a high proportion of outliers.

\begin{table}
\footnotesize
\caption{Pearson correlation coefficients between the proportion of minority objects of a given type and the rank of the RBU algorithm with regard to the given metric. Statistically significant correlations denoted with bold font. Note that negative correlation indicates increasing performance.}
\label{table:type}
\centering
\begin{tabularx}{\linewidth}{llllll}
\toprule
& Metric & Safe & Borderline & Rare & Outlier \\
\midrule
\parbox[t]{2mm}{\multirow{5}{*}{\rotatebox[origin=c]{90}{CART}}} & Precision & \textbf{-0.3770} & -0.0434 & \textbf{+0.3329} & +0.2395 \\
 & Recall & \textbf{+0.6517} & +0.1797 & \textbf{-0.4036} & \textbf{-0.5895} \\
 & F-measure & +0.0792 & -0.1179 & -0.1292 & +0.0913 \\
 & AUC & \textbf{+0.3760} & +0.0493 & -0.1294 & \textbf{-0.3414} \\
 & G-mean & \textbf{+0.4637} & +0.0632 & -0.2387 & \textbf{-0.3852} \\
\midrule
\parbox[t]{2mm}{\multirow{5}{*}{\rotatebox[origin=c]{90}{KNN}}} & Precision & -0.2934 & -0.1719 & \textbf{+0.4244} & +0.2299 \\
 & Recall & \textbf{+0.6305} & \textbf{+0.4247} & \textbf{-0.5032} & \textbf{-0.7397} \\
 & F-measure & +0.0180 & -0.0536 & +0.1097 & -0.0221 \\
 & AUC & \textbf{+0.4862} & +0.1406 & -0.3007 & \textbf{-0.4458} \\
 & G-mean & \textbf{+0.5472} & +0.3050 & \textbf{-0.3676} & \textbf{-0.6185} \\
\midrule
\parbox[t]{2mm}{\multirow{5}{*}{\rotatebox[origin=c]{90}{NB}}} & Precision & -0.1286 & +0.0021 & -0.0371 & +0.1389 \\
 & Recall & -0.1524 & -0.2858 & \textbf{+0.4037} & +0.2064 \\
 & F-measure & -0.1802 & -0.0823 & +0.0476 & +0.2227 \\
 & AUC & -0.2582 & \textbf{-0.4682} & +0.1975 & \textbf{+0.5690} \\
 & G-mean & -0.2204 & -0.2354 & +0.2270 & \textbf{+0.3114} \\
\midrule
\parbox[t]{2mm}{\multirow{5}{*}{\rotatebox[origin=c]{90}{SVM}}} & Precision & -0.0207 & +0.0583 & +0.0107 & -0.0373 \\
 & Recall & \textbf{+0.4724} & \textbf{+0.4331} & \textbf{-0.4653} & \textbf{-0.6141} \\
 & F-measure & -0.0799 & +0.2325 & +0.0224 & -0.1414 \\
 & AUC & +0.0024 & -0.0350 & -0.1779 & +0.1143 \\
 & G-mean & +0.2390 & +0.2559 & -0.2864 & \textbf{-0.3189} \\
\bottomrule
\end{tabularx}
\end{table}

\section{Conclusions}
\label{sec:con}

Throughout this paper we proposed a novel undersampling algorithm, Radial-Based Undersampling, based on a previously introduced concept of mutual class potential. The main motivation behind the proposed algorithm was extending the notion of non-nearest neighbor based resampling, previously used in Radial-Based Oversampling, to the undersampling procedure. The proposed method offers a conceptually simple and computationally more efficient alternative to the Radial-Based Oversampling algorithm. In the conducted experimental study we empirically evaluated the usefulness of the proposed method. Through the course of the study we were able to identify the areas of applicability of the algorithm. Specifically, the observed results indicate the suitability of the algorithm to be used in combination with naive Bayes classifier and, to a lesser extent, CART decision tree and support vector machine. Compared to the Radial-Based Oversampling, RBU displayed a statistically significantly better performance when combined with CART decision tree. Furthermore, we were able to analyse the behavior of the proposed algorithm with respect to the characteristics of datasets on which it was applied. For the majority of the examined classification algorithms proposed method achieved comparatively better results when used on difficult datasets, consisting of higher proportion of rare and outlier minority instances.

Despite the relative simplicity of the proposed criterion of undersampling selection, the observed results are encouraging for further development of the algorithm. Specifically, we intend to explore the possibility of using other selection criteria based on the idea of mutual class potential, with a particular focus on more theoretically motivated choices.

\section*{Acknowledgments}

This work was supported by the Polish National Science Center under the grant no. 2017/27/N/ST6/01705 as well as the PLGrid Infrastructure.


\bibliographystyle{elsarticle-num}
\bibliography{main}

\begin{thebibliography}{10}
\expandafter\ifx\csname url\endcsname\relax
  \def\url#1{\texttt{#1}}\fi
\expandafter\ifx\csname urlprefix\endcsname\relax\def\urlprefix{URL }\fi
\expandafter\ifx\csname href\endcsname\relax
  \def\href#1#2{#2} \def\path#1{#1}\fi

\bibitem{Sun:2009}
Y.~Sun, A.~K.~C. Wong, M.~S. Kamel, Classification of imbalanced data: A
  review, International Journal of Pattern Recognition and Artificial
  Intelligence 23~(4) (2009) 687--719.

\bibitem{Krawczyk:2016}
B.~Krawczyk, Learning from imbalanced data: open challenges and future
  directions, Progress in Artificial Intelligence 5~(4) (2016) 221--232.

\bibitem{Branco:2016}
P.~Branco, L.~Torgo, R.~P. Ribeiro, A survey of predictive modeling on
  imbalanced domains, ACM Computing Surveys 49~(2) (2016) 31:1--31:50.

\bibitem{jo2004class}
T.~Jo, N.~Japkowicz, Class imbalances versus small disjuncts, ACM Sigkdd
  Explorations Newsletter 6~(1) (2004) 40--49.

\bibitem{chen2008fast}
X.-w. Chen, M.~Wasikowski, Fast: a {ROS}-based feature selection metric for
  small samples and imbalanced data classification problems, in: Proceedings of
  the 14th ACM SIGKDD international conference on Knowledge discovery and data
  mining, ACM, 2008, pp. 124--132.

\bibitem{krawczyk2016evolutionary}
B.~Krawczyk, M.~Galar, {\L}.~Jele{\'n}, F.~Herrera, Evolutionary undersampling
  boosting for imbalanced classification of breast cancer malignancy, Applied
  Soft Computing 38 (2016) 714--726.

\bibitem{koziarski2018convolutional}
M.~Koziarski, B.~Kwolek, B.~Cyganek, Convolutional neural network-based
  classification of histopathological images affected by data imbalance, in:
  Video Analytics. Face and Facial Expression Recognition, Springer, 2018, pp.
  1--11.

\bibitem{ramentol2016fuzzy}
E.~Ramentol, I.~Gondres, S.~Lajes, R.~Bello, Y.~Caballero, C.~Cornelis,
  F.~Herrera, Fuzzy-rough imbalanced learning for the diagnosis of high voltage
  circuit breaker maintenance: The {SMOTE-FRST-2T} algorithm, Engineering
  Applications of Artificial Intelligence 48 (2016) 134--139.

\bibitem{wei2013effective}
W.~Wei, J.~Li, L.~Cao, Y.~Ou, J.~Chen, Effective detection of sophisticated
  online banking fraud on extremely imbalanced data, World Wide Web 16~(4)
  (2013) 449--475.

\bibitem{azaria2014behavioral}
A.~Azaria, A.~Richardson, S.~Kraus, V.~Subrahmanian, Behavioral analysis of
  insider threat: A survey and bootstrapped prediction in imbalanced data, IEEE
  Transactions on Computational Social Systems 1~(2) (2014) 135--155.

\bibitem{czarnecki2015compounds}
W.~M. Czarnecki, K.~Rataj, Compounds activity prediction in large imbalanced
  datasets with substructural relations fingerprint and {EEM}, in: 2015 IEEE
  Trustcom/BigDataSE/ISPA, Vol.~2, IEEE, 2015, pp. 192--192.

\bibitem{fernandez2017pareto}
A.~Fern{\'a}ndez, C.~J. Carmona, M.~Jose~del Jesus, F.~Herrera, A
  {P}areto-based ensemble with feature and instance selection for learning from
  multi-class imbalanced datasets, International Journal of neural systems
  27~(06) (2017) 1750028.

\bibitem{koziarski2017ccr}
M.~Koziarski, M.~Wo{\.z}niak, {CCR}: A combined cleaning and resampling
  algorithm for imbalanced data classification, International Journal of
  Applied Mathematics and Computer Science 27~(4) (2017) 727--736.

\bibitem{lango2018multi}
M.~Lango, J.~Stefanowski, Multi-class and feature selection extensions of
  roughly balanced bagging for imbalanced data, Journal of Intelligent
  Information Systems 50~(1) (2018) 97--127.

\bibitem{ksieniewicz2018imbalanced}
P.~Ksieniewicz, M.~Wo{\'z}niak, Imbalanced data classification based on feature
  selection techniques, in: International Conference on Intelligent Data
  Engineering and Automated Learning, Springer, 2018, pp. 296--303.

\bibitem{chawla2002smote}
N.~V. Chawla, K.~W. Bowyer, L.~O. Hall, W.~P. Kegelmeyer, {SMOTE}: synthetic
  minority over-sampling technique, Journal of Artificial Intelligence Research
  16 (2002) 321--357.

\bibitem{koziarski2017radial}
M.~Koziarski, B.~Krawczyk, M.~Wo{\'z}niak, Radial-based approach to imbalanced
  data oversampling, in: International Conference on Hybrid Artificial
  Intelligence Systems, Springer, 2017, pp. 318--327.

\bibitem{perez2016oversampling}
M.~P{\'e}rez-Ortiz, P.~A. Guti{\'e}rrez, P.~Tino, C.~Herv{\'a}s-Mart{\'\i}nez,
  Oversampling the minority class in the feature space, IEEE transactions on
  neural networks and learning systems 27~(9) (2016) 1947--1961.

\bibitem{bellinger2018manifold}
C.~Bellinger, C.~Drummond, N.~Japkowicz, Manifold-based synthetic oversampling
  with manifold conformance estimation, Machine Learning 107~(3) (2018)
  605--637.

\bibitem{han2005borderline}
H.~Han, W.-Y. Wang, B.-H. Mao, Borderline-{SMOTE}: a new over-sampling method
  in imbalanced data sets learning, in: International Conference on Intelligent
  Computing, Springer, 2005, pp. 878--887.

\bibitem{he2008adasyn}
H.~He, Y.~Bai, E.~A. Garcia, S.~Li, {ADASYN}: Adaptive synthetic sampling
  approach for imbalanced learning, in: 2008 IEEE International Joint
  Conference on Neural Networks (IEEE World Congress on Computational
  Intelligence), IEEE, 2008, pp. 1322--1328.

\bibitem{Bunkhumpornpat:2009}
C.~Bunkhumpornpat, K.~Sinapiromsaran, C.~Lursinsap, {Safe-Level-SMOTE}:
  safe-level-synthetic minority over-sampling technique for handling the class
  imbalanced problem, in: Advances in Knowledge Discovery and Data Mining, 13th
  Pacific-Asia Conference 2009, Bangkok, Thailand, April 27-30, 2009,
  Proceedings, 2009, pp. 475--482.

\bibitem{Maciejewski:2011}
T.~Maciejewski, J.~Stefanowski, Local neighbourhood extension of {SMOTE} for
  mining imbalanced data, in: Proceedings of the {IEEE} Symposium on
  Computational Intelligence and Data Mining 2011, part of the {IEEE} Symposium
  Series on Computational Intelligence 2011, April 11-15, 2011, Paris, France,
  2011, pp. 104--111.

\bibitem{tomek1976two}
I.~Tomek, Two modifications of {CNN}, IEEE Transactions on Systems, Man, and
  Cybernetics 6 (1976) 769--772.

\bibitem{wilson1972asymptotic}
D.~L. Wilson, Asymptotic properties of nearest neighbor rules using edited
  data, IEEE Transactions on Systems, Man, and Cybernetics 2~(3) (1972)
  408--421.

\bibitem{hart1968condensed}
P.~Hart, The condensed nearest neighbor rule, IEEE transactions on information
  theory 14~(3) (1968) 515--516.

\bibitem{mani2003knn}
I.~Mani, I.~Zhang, k{NN} approach to unbalanced data distributions: a case
  study involving information extraction, in: Proceedings of workshop on
  learning from imbalanced datasets, Vol. 126, 2003.

\bibitem{anand2010approach}
A.~Anand, G.~Pugalenthi, G.~B. Fogel, P.~Suganthan, An approach for
  classification of highly imbalanced data using weighting and undersampling,
  Amino acids 39~(5) (2010) 1385--1391.

\bibitem{smith2014instance}
M.~R. Smith, T.~Martinez, C.~Giraud-Carrier, An instance level analysis of data
  complexity, Machine learning 95~(2) (2014) 225--256.

\bibitem{yen2009cluster}
S.-J. Yen, Y.-S. Lee, Cluster-based under-sampling approaches for imbalanced
  data distributions, Expert Systems with Applications 36~(3) (2009)
  5718--5727.

\bibitem{liu2008exploratory}
X.-Y. Liu, J.~Wu, Z.-H. Zhou, Exploratory undersampling for class-imbalance
  learning, IEEE Transactions on Systems, Man, and Cybernetics, Part B
  (Cybernetics) 39~(2) (2008) 539--550.

\bibitem{galar2013eusboost}
M.~Galar, A.~Fern{\'a}ndez, E.~Barrenechea, F.~Herrera, {EUSB}oost: Enhancing
  ensembles for highly imbalanced data-sets by evolutionary undersampling,
  Pattern Recognition 46~(12) (2013) 3460--3471.

\bibitem{lu2017adaptive}
W.~Lu, Z.~Li, J.~Chu, Adaptive ensemble undersampling-boost: a novel learning
  framework for imbalanced data, Journal of Systems and Software 132 (2017)
  272--282.

\bibitem{barandela2004imbalanced}
R.~Barandela, R.~M. Valdovinos, J.~S. S{\'a}nchez, F.~J. Ferri, The imbalanced
  training sample problem: Under or over sampling?, in: Joint IAPR
  international workshops on statistical techniques in pattern recognition
  (SPR) and structural and syntactic pattern recognition (SSPR), Springer,
  2004, pp. 806--814.

\bibitem{drummond2003c4}
C.~Drummond, R.~C. Holte, et~al., {C4.5}, class imbalance, and cost
  sensitivity: why under-sampling beats over-sampling, in: Workshop on learning
  from imbalanced datasets II, Vol.~11, Citeseer, 2003, pp. 1--8.

\bibitem{van2009knowledge}
J.~Van~Hulse, T.~Khoshgoftaar, Knowledge discovery from imbalanced and noisy
  data, Data \& Knowledge Engineering 68~(12) (2009) 1513--1542.

\bibitem{garcia2012effectiveness}
V.~Garc{\'\i}a, J.~S. S{\'a}nchez, R.~A. Mollineda, On the effectiveness of
  preprocessing methods when dealing with different levels of class imbalance,
  Knowledge-Based Systems 25~(1) (2012) 13--21.

\bibitem{wolpert1996lack}
D.~H. Wolpert, The lack of a priori distinctions between learning algorithms,
  Neural computation 8~(7) (1996) 1341--1390.

\bibitem{napierala2016types}
K.~Napierala, J.~Stefanowski, Types of minority class examples and their
  influence on learning classifiers from imbalanced data, Journal of
  Intelligent Information Systems 46~(3) (2016) 563--597.

\bibitem{koziarski2019radial}
M.~Koziarski, B.~Krawczyk, M.~Wo{\'z}niak, Radial-{B}ased {O}versampling for
  noisy imbalanced data classification, Neurocomputing.

\bibitem{koziarski2019mcrbo}
B.~Krawczyk, M.~Koziarski, M.~Wo{\'z}niak, Radial-{B}ased {O}versampling for
  multi-class imbalanced data classification, IEEE Transactions on Neural
  Networks and Learning Systems.

\bibitem{bobowska2018experimental}
B.~Bobowska, M.~Wo{\'z}niak, Experimental study on {M}odified {R}adial-{B}ased
  {O}versampling, in: The 13th International Conference on Soft Computing
  Models in Industrial and Environmental Applications, Springer, 2018, pp.
  110--119.

\bibitem{alcala2011keel}
J.~Alcal{\'a}-Fdez, A.~Fern{\'a}ndez, J.~Luengo, J.~Derrac, S.~Garc{\'\i}a,
  L.~S{\'a}nchez, F.~Herrera, {KEEL} data-mining software tool: data set
  repository, integration of algorithms and experimental analysis framework.,
  Journal of Multiple-Valued Logic \& Soft Computing 17.

\bibitem{pedregosa2011scikit}
F.~Pedregosa, G.~Varoquaux, A.~Gramfort, V.~Michel, B.~Thirion, O.~Grisel,
  M.~Blondel, P.~Prettenhofer, R.~Weiss, V.~Dubourg, et~al., Scikit-learn:
  Machine learning in {P}ython, Journal of Machine Learning Research 12~(Oct)
  (2011) 2825--2830.

\bibitem{tomek1976experiment}
I.~Tomek, An experiment with the edited nearest-neighbor rule, IEEE
  Transactions on systems, Man, and Cybernetics~(6) (1976) 448--452.

\bibitem{laurikkala2001improving}
J.~Laurikkala, Improving identification of difficult small classes by balancing
  class distribution, in: Conference on Artificial Intelligence in Medicine in
  Europe, Springer, 2001, pp. 63--66.

\bibitem{Kubat:1997}
M.~Kubat, S.~Matwin, Addressing the curse of imbalanced training sets:
  One-sided selection, in: In Proceedings of the 14th International Conference
  on Machine Learning, Morgan Kaufmann, 1997, pp. 179--186.

\bibitem{lemaitre2017imbalanced}
G.~Lemaitre, F.~Nogueira, C.~K. Aridas, Imbalanced-learn: A {P}ython toolbox to
  tackle the curse of imbalanced datasets in machine learning, Journal of
  Machine Learning Research 18~(17) (2017) 1--5.

\bibitem{alpaydin1999combined}
E.~Alpaydin, Combined 5 $\times$ 2 cv {F} test for comparing supervised
  classification learning algorithms, Neural Computation 11~(8) (1999)
  1885--1892.

\end{thebibliography}

\end{document}